\documentclass[10pt,twocolumn,letterpaper]{article}

\usepackage[pagenumbers]{cvpr}

\usepackage{amsmath,amssymb}
\usepackage{booktabs}
\usepackage{colortbl}
\definecolor{tintred}{HTML}{F6D7D9}
\definecolor{tintblue}{HTML}{DAE8FC}
\usepackage{graphicx}
\usepackage{algorithm}
\usepackage{algpseudocode}

\usepackage[T1,OT1]{fontenc}
\usepackage[skins,breakable,listings]{tcolorbox}
\definecolor{PromptBorder}{HTML}{065F5C}
\definecolor{PromptBack}{HTML}{F2F9F8}
\newcommand{\promptfont}{\fontencoding{T1}\fontfamily{DejaVuSansMono-TLF}\fontsize{8.5}{10.2}\selectfont}
\newtcblisting{promptbox}[2][]{
  enhanced,
  unbreakable,
  colframe=PromptBorder,
  colback=PromptBack,
  colbacktitle=PromptBorder,
  coltitle=white,
  fonttitle=\bfseries,
  title={#2},
  arc=3pt,
  boxrule=1.5pt,
  left=6pt,right=6pt,top=4pt,bottom=4pt,
  before skip=6pt,after skip=6pt,
  listing only,
  listing options={
    basicstyle=\promptfont,
    breaklines=true,
    breakautoindent=false,
    breakindent=0pt,
    breakatwhitespace=true,
    columns=fullflexible,
    keepspaces=true,
    showstringspaces=false
  },
  #1
}

\usepackage{capt-of}

\definecolor{cvprblue}{rgb}{0.21,0.49,0.74}
\usepackage{colortbl}
\usepackage{tabularx}
\definecolor{lighterrow}{RGB}{240,247,245}
\definecolor{oursrow}{RGB}{230,242,239}
\usepackage[pagebackref,breaklinks,colorlinks,allcolors=cvprblue]{hyperref}
\usepackage{array}
\usepackage{booktabs}
\usepackage{multirow}

\AtBeginDocument{%
  \setlength{\abovedisplayskip}{4pt plus 2pt minus 2pt}%
  \setlength{\belowdisplayskip}{4pt plus 2pt minus 2pt}%
  \setlength{\abovedisplayshortskip}{4pt plus 2pt minus 2pt}%
  \setlength{\belowdisplayshortskip}{4pt plus 2pt minus 2pt}%
}

\def\confName{CVPR}
\def\confYear{2027}

\title{ ReFocus: Progress-Guided Adaptive Practice for Long-Horizon Robotic Manipulation Reinforcement Learning}

\author{Haoyuan Li$^1$ \quad Wei Huang$^2$ \quad Jiehong Lin$^2$ \quad Shengrong Zhang$^2$ \quad Chenzhi Liu$^2$ \quad Shengchao Zhou$^2$ \\
Hehe Fan$^1$ \quad Xiaojuan Qi$^{2,\dagger}$ \quad Yi Yang$^{1,\dagger}$ \\
{$^1$ Zhejiang University \quad $^2$ The University of Hong Kong} 
}

\begin{document}
\maketitle
\begingroup
\renewcommand{\thefootnote}{\fnsymbol{footnote}}
\footnotetext[2]{Corresponding authors.}
\endgroup

\begin{abstract}
Online reinforcement learning (RL) enables vision-language-action policies to improve through environmental interaction. However, in long-horizon manipulation, sparse success/failure feedback provides little information about where policies encounter bottlenecks. Moreover, repeated full-task rollouts also leave these bottlenecks weakly analyzed and insufficiently practiced. To this end, we introduce \textbf{ReFocus}, a progress-guided multi-round RL framework that alternates Global with Local RL to identify task-specific bottlenecks and allocate additional practice to them. Specifically, during Global RL, we decompose each task into a Directed Acyclic Graph (DAG) of subtasks. Then a DAG-based Progress Reward is leveraged for both policy training and bottleneck diagnosis. Based on the identified bottlenecks, ReFocus selects Adaptive Practice Windows, defining task segments from an entry stage to a target stage. It allocates additional interaction to these segments during Local RL while updating the entire policy. After that, it evaluates the policy to perform the full-task for the next round RL, forming an iteration between Global and Local RL. Experiments on LIBERO-Long and RoboCasa365 show that ReFocus achieves success rates of 97.8\% and 51.3\% respectively, compared with 90.6\% and 39.3\% for the $\pi_{0.5}$ baseline. Furthermore, we evaluate a real-world adaptation on three long-horizon tasks. On BreadBaker, ReFocus achieves 95\% success, compared with 30\% for the $\pi_{0.5}$. Project page: {\urlstyle{same}\url{https://refocus-robotics.github.io/}}
\end{abstract}
    
\section{Introduction}
\label{sec:intro}

In long-horizon manipulation, the success of a subtask depends on both its execution and the state inherited from preceding stages. Completing one subtask therefore does not guarantee successful continuation. Imitation learning gives vision-language-action (VLA) policies a strong initialization by supervising actions along expert trajectories~\cite{black2025pi0,black2025pi05,bjorck2025gr00t}. At deployment, however, successive subtasks begin from states produced by the policy itself, and demonstrations provide limited supervision for handling the deviations that accumulate along the way, resulting in failure.

\begin{figure*}[t]
    \centering
    \includegraphics[width=\textwidth]{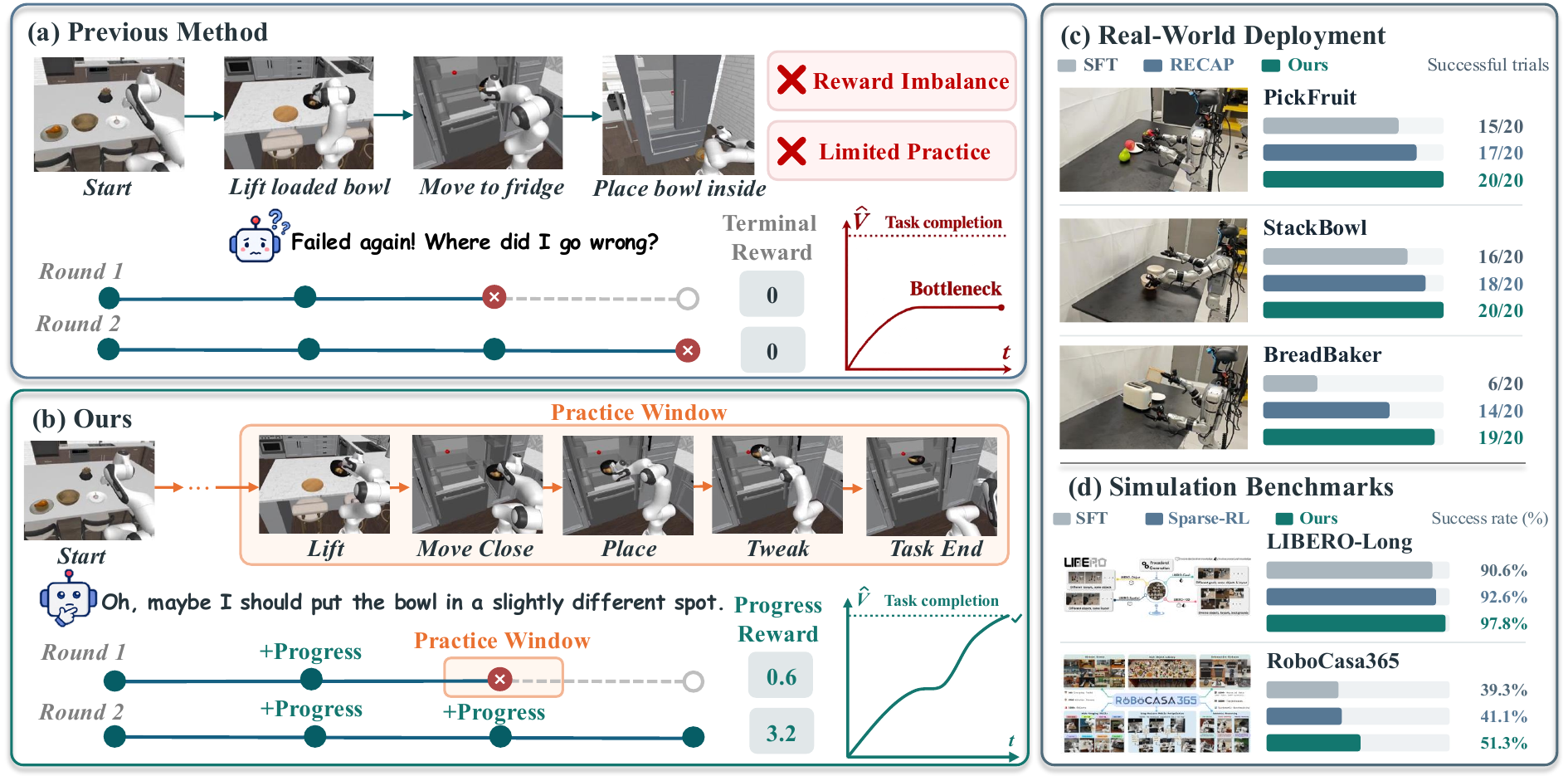}
    \caption{\textbf{Motivation and overview of ReFocus.} \textbf{(a)} Sparse terminal rewards reveal little about partial progress or task-specific bottlenecks. \textbf{(b)} ReFocus uses progress feedback to guide targeted practice through alternation of Global RL and Local RL. \textbf{(c)} ReFocus achieves success rates of 98.3\% on three real-world long-horizon tasks, compared with 81.7\% for RECAP~\cite{physicalintelligence2025recap}. \textbf{(d)} ReFocus raises the success rates of $\pi_{0.5}$~\cite{black2025pi05} from 90.6\% to 97.8\% on LIBERO-Long and from 39.3\% to 51.3\% on RoboCasa365.}
    \label{fig:repractice_motivation}
    \vspace{-10pt}
\end{figure*}

Online reinforcement learning (RL) enables VLA policies to learn beyond demonstrations with further environmental interactions by themselves. However, most online RL methods, which learn from full-task rollouts using only binary terminal feedback, still struggle with long-horizon tasks.  As illustrated in Fig.~\ref{fig:repractice_motivation}(a), two episodes with different failures—one from bowl transport and the other from shelf placement—receive the same zero reward despite different amounts of completed work. Our diagnostic further reveals concentrations of failures at specific stages, as shown in Figs.~\ref{fig:robocasa_failure_stages} and~\ref{fig:robocasa_task_bottlenecks} of Appendix~\ref{sec:supp_failure_stages}.

Although some work recovers intermediate supervision by estimating progress between observations or rewarding verifiable subtask completions~\cite{tan2026general,yin2026structrl}, two challenges remain: \textbf{1) Reward Imbalance.} To characterize intermediate progress, we group verifiable subtasks with shared prerequisites into stages, within which subtasks can be completed in any order. Even with prerequisite gating and one-time credit, the total reward assigned to a stage can depend on its decomposition granularity. Uniform per-subtask rewards, for example, give a stage more weight simply because it contains more completion events. \textbf{2) Limited Bottleneck Practice.} Under a fixed interaction budget, the stage limiting full-task success can receive few opportunities for practice and less comprehensive analysis. For example, in Fig.~\ref{fig:repractice_motivation}(a), each new attempt at shelf placement requires another grasp and transport, and an earlier failure prevents that attempt altogether. Moreover, there is no deeper analysis of the reason for the failure of shelf placement.

To this end, we introduce \textbf{ReFocus}, a progress-guided RL framework for long-horizon manipulation. ReFocus alternates Global RL with Local RL to improve a shared VLA policy. During Global RL, we represent each task as a directed acyclic graph (DAG) of subtasks and their dependencies. To address Reward Imbalance, DAG-based Progress Reward allocates a fixed task-level reward budget equally across stages. Each stage's share is then divided equally among its subtasks. The total budget allocated to a stage is therefore independent of its number of subtasks, preventing finer decompositions from receiving disproportionate weight. Subtask-completion records also provide structured evidence of task progress to support bottleneck diagnosis.

Based on progress feedback, an LLM-based Scheduler identifies stages where the policy repeatedly stalls and selects Adaptive Practice Windows for Local RL. \emph{Normal window} rollouts concentrate practice on the selected bottleneck, while \emph{prefix window} and \emph{suffix window} rollouts connect it with execution from the normal task start and continuation toward full-task completion, respectively. Milestone rewards provide finer feedback within the selected task segment. Together, these rollout modes connect targeted practice with the states inherited from earlier stages and the requirements of later stages. After each round, evaluations assess task progress and full-task success. The Scheduler uses these results, previous windows, and interaction costs to retain, revise, or skip subsequent practice windows. By keeping practice allocation aligned with the remaining difficulties, multi-round training helps translate local gains into progressively more reliable full-task execution.

We evaluate ReFocus on LIBERO-Long~\cite{liu2023libero} and RoboCasa365~\cite{robocasa365}. Based on a supervised finetuning (SFT) baseline $\pi_{0.5}$~\cite{black2025pi05}, ReFocus improves its performance from 90.6\% and 39.3\% to 97.8\% and 51.3\% respectively. On three long-horizon real-world tasks, we compare a RECAP-based adaptation of ReFocus with SFT and RECAP~\cite{physicalintelligence2025recap}. On \texttt{BreadBaker}, RECAP improves the success rate from 30\% to 70\%. ReFocus further increases it to 95\%. ReFocus also achieves 100\% on both \texttt{StackBowl} and \texttt{PickFruit}, compared with 90\% and 85\% for RECAP, respectively.

\begin{itemize}
   \item We introduce ReFocus, a progress-guided RL framework that alternates Global RL with Local RL to identify task-specific bottlenecks and allocate targeted practice in long-horizon manipulation.
   \item ReFocus uses DAG-based progress rewards during Global RL to provide dense feedback and identify bottleneck subtasks. Adaptive Practice Windows then focus Local RL on these bottlenecks.
   \item Extensive experiments show that ReFocus improves the long-horizon manipulation performance of SFT baselines on the LIBERO-Long and RoboCasa365 benchmarks and on real-world manipulation tasks.
\end{itemize}

\begin{figure*}[t]
    \centering
    \includegraphics[width=\textwidth]{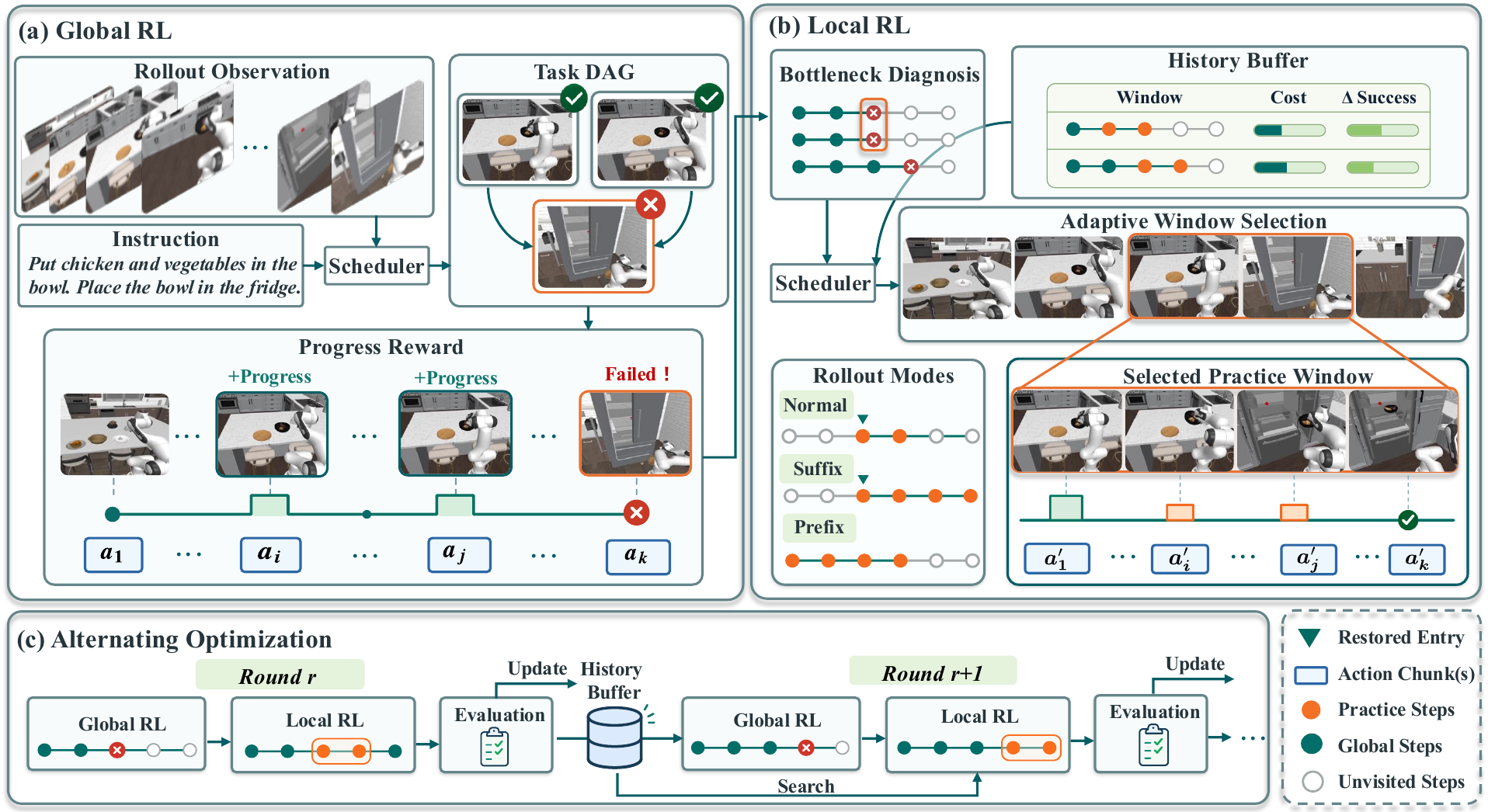}
    \caption{\textbf{Illustration of ReFocus.} \textbf{(a)} Global RL uses task DAGs to assign progress rewards and track subtask completion. \textbf{(b)} The Scheduler selects Adaptive Practice Windows for bottleneck practice. Three rollout modes connect local practice with full-task execution, while milestone rewards provide finer feedback. \textbf{(c)} Multi-round training alternates Global and Local RL. Full-task feedback and practice costs are recorded in the history buffer to guide subsequent practice.}
    \label{fig:repractice_method}
    \vspace{-10pt}
\end{figure*}
\section{Related Work}
\label{sec:related_work}

\noindent\textbf{Robotic Foundation Models.} Large-scale robot demonstrations enable VLA models to learn generalist manipulation across diverse tasks and embodiments~\cite{black2025pi0,black2025pi05,bjorck2025gr00t,kim2025openvla}. Recent efforts further scale heterogeneous robot data to improve policy capabilities~\cite{wu2026lingbotvla2,xiaomi2026robotics1,meituan2026robotics0,yuan2026qwenrobotmanip,zhang2026hyvla}. Embodied reasoning and steerable conditioning support multi-step execution~\cite{gemini2025robotics15,physicalintelligence2026pi07}, while world-action models integrate visual prediction with robot control~\cite{ye2026dreamzero,nvidia2026gr00t2,huang2026long}. However, policies trained on demonstrations may struggle with unfamiliar states and accumulated errors during long-horizon manipulation, motivating further improvement through RL.

\noindent\textbf{Reinforcement Learning for VLAs.} Reinforcement learning improves pretrained VLA policies using offline data or online interaction. Offline methods use preferences, action values, or recovery data~\cite{zhang2024grape,huang2025corft,gao2026ogpo}, while online methods optimize policies with actor--critic or group-relative objectives~\cite{lu2025vlarl,liu2025rlvlageneralization,li2026simplevla,chen2025tgrpo}. Further work studies efficient training and real-world adaptation~\cite{zang2026rlinfvla,wang2026policytrim,chen2025conrft,physicalintelligence2025recap,wagenmaker2025dsrl,dong2026expoft}. For long-horizon tasks, terminal outcomes provide limited feedback on partially completed trajectories. To provide feedback beyond task outcomes, reward and critic models incorporate intermediate progress~\cite{zhang2026vlac,tan2026general,liang2026robometer,fei2025srpo,kim2026progvla}, stage structure~\cite{chen2026sarm2,feng2026procvlm}, or execution history~\cite{xu2026robodopamine2,chen2026robolongr}. Building on these approaches, ReFocus uses DAG-based progress feedback to allocate equal reward budgets across stages and credit subtask completions according to prerequisites.

\noindent\textbf{Long-Horizon Manipulation.} Long-horizon tasks require policies to complete multiple subtasks while handling errors accumulated during execution. Prior work addresses long-horizon tasks by incorporating subtask prerequisites or downstream success into rewards~\cite{yin2026structrl,liu2026foresight}. Other approaches focus learning on difficult subtasks through specialist policies~\cite{su2026parts,jiang2026repair,liu2026mobiagent}, or allocate practice using budget-aware scheduling, reset curricula, and task selection~\cite{vats2026deliberate,florensa2017reverse,aphale2026percontext,fang2026find}. Execution feedback has also been used to improve skills and policies across iterations~\cite{wang2026adarsi,chen2026recursia,xu2026baton}. In contrast, ReFocus adaptively selects task-specific practice windows across rounds using subtask progress, full-task evaluations, and practice history.

\section{Method}
\label{sec:method}

In this section, we first introduce Global RL with Progress Reward for full-task policy learning (\textbf{\S}\ref{sec:global_rl}). We then present Local RL, which directs interaction toward stages requiring further improvement (\textbf{\S}\ref{sec:window_selection}). Finally, we describe the optimization paradigm that alternates between phases to improve a shared policy across training rounds (\textbf{\S}\ref{sec:global_local}). The overall pipeline of ReFocus is shown in Fig.~\ref{fig:repractice_method}.

\subsection{Global RL with Progress Reward}
\label{sec:global_rl}

We consider a set of long-horizon manipulation tasks $\mathcal T$ and a shared VLA policy $\pi_\theta$ initialized through supervised fine-tuning (SFT). At each decision step $k$, the policy receives an input $o_k$, comprising the visual observation and task instruction, and generates an action chunk $a_k$. Global RL optimizes this policy using full-task rollouts. To provide denser feedback than terminal rewards, we use a frontier-LLM-driven Scheduler $\mathcal{S}$ to construct task DAGs and define progress rewards for valid subtask completions.

\noindent\textbf{Task DAG.}
Each task is represented by a directed acyclic graph $G=(V,E)$, where nodes $V$ denote subtasks and edges $E$ specify prerequisites. Subtasks with the same prerequisites form an ordered stage and can be completed in any order. For example, placing two vegetables in a basin forms a stage with two subtasks. Both placements must be credited before the next stage becomes eligible for reward.

\noindent\textbf{DAG-based Progress Reward.}
To balance progress rewards across stages, we allocate an equal reward budget to each stage and divide it uniformly among its subtasks. Let $z_{v,k}\in\{0,1\}$ indicate whether subtask $v$ receives its first valid completion credit at action chunk $k$, after its prerequisites have been credited. The accumulated credits define the progress state $\boldsymbol{p}_k=(p_{v,k})_{v\in V}$, where $p_{v,k}=\sum\nolimits_{j=0}^{k}z_{v,j}$. For a task with $S$ stages, we distribute a total progress reward budget of 2, giving $B_s=2/S$. Stage $s$ contains the subtask set $V_s$, with $n_s=|V_s|$. Each subtask receives a reward of $B_s/n_s$. For a fixed stage partition, subdividing a stage into more subtasks leaves its total progress-reward budget unchanged. Therefore, the progress reward for action chunk $k$ can be summarised as:
\begin{equation}
r^{\mathrm{prog}}_k =
\sum_{s=1}^{S}
\frac{B_s}{n_s}
\sum_{v\in V_s}z_{v,k}.
\label{eq:progress_reward}
\end{equation}

The total reward combines progress feedback with terminal reward:
\begin{equation}
r_k^{\mathrm G}
=
r^{\mathrm{prog}}_k
+
\lambda_{\mathrm{succ}}z^{\mathrm{succ}}_k,
\label{eq:repractice_reward}
\end{equation}
where $z^{\mathrm{succ}}_k$ indicates the full-task success and $\lambda_{\mathrm{succ}}$ is its reward weight. Global RL uses $r_k^{\mathrm G}$ to optimize full-task execution, while the recorded progress states support subsequent bottleneck diagnosis and practice-window selection.

\subsection{Local RL with Adaptive Practice}
\label{sec:window_selection}

Local RL focuses interaction on bottleneck stages that remain difficult for the current policy. The Scheduler $\mathcal S$ uses validation rollouts and the History Buffer $\mathcal H$ to select task-specific practice windows and specify fine-grained milestone rewards for updating the shared policy.

\noindent\textbf{Bottleneck Diagnosis.} To identify stages that require additional practice, we evaluate the policy on the validation set every five updates during Global RL. Each validation rollout provides a progress sequence $\boldsymbol{p}_{0:K}$, an observation sequence $o_{0:K}$, and the full-task outcome, where $K$ denotes the final decision step. We summarize these records as $\mathcal D$. The Scheduler $\mathcal S$ compares progress sequences across unsuccessful rollouts to identify the earliest stage where progress repeatedly stalls. It then examines the corresponding visual observations to describe the stalled actions and supporting evidence, reporting uncertainty when the evidence is insufficient. The resulting diagnosis guides practice-window selection.

\noindent\textbf{Adaptive Window Selection.} The Scheduler combines the current diagnosis $\mathcal D$ with the History Buffer $\mathcal H$, which stores information from previous practice rounds, to select a practice window around the identified bottleneck. This information guides whether to retain the previous window, revise its scope, or skip further practice. The Scheduler specifies a window $w=(s_{\mathrm{entry}},s_{\mathrm{target}},B)$, where $s_{\mathrm{entry}}$ defines the starting progress, $s_{\mathrm{target}}$ defines the stage to complete, and $B$ limits the environment interaction within the window. Along with the window, it provides a set of milestones $\mathcal M$ and their verifiable completion conditions to guide practice within the selected task segment. For example, shelf placement can be decomposed into secure holding, alignment, and placement with release. The Scheduler also retrieves a training entry bank $\mathcal B$ containing available restorable states that satisfy the entry condition:
\begin{equation}
(w,\mathcal B,\mathcal M)=
\mathcal S(\mathcal D,\mathcal H;\mathcal E),
\label{eq:window_selection}
\end{equation}
where $\mathcal E$ represents the available training entries. We set $w=\varnothing$ when no practice window is selected.

\noindent\textbf{Rollout Modes and Milestone Rewards.} During Local RL, we sample one of three rollout modes with probabilities $p_m$, where $m\in\{\mathrm{norm},\mathrm{pre},\mathrm{suf}\}$ and $\sum_m p_m=1$. \emph{Normal window} rollouts start from restored entries and focus on the selected bottleneck, ending at target completion or budget exhaustion. \emph{Prefix window} rollouts start from normal task states and continue to the window target, while \emph{Suffix window} rollouts start from restored entries and continue toward full-task completion within the remaining task horizon. Together, these modes connect bottleneck practice with preceding and subsequent task execution. Tasks without an eligible window use normal-start full-task rollouts. For task $i$, let $v$ denote the refined target subtask and $s$ its stage. The Local RL reward can be summarized as:
\begin{equation} r_k^{\mathrm L} =r_k^{\mathrm{prog}}-\frac{B_s}{n_s}z_{v,k} +\sum\nolimits_{j\in\mathcal M_i}\eta_j z^{\mathrm{mile}}_{j,k} +\lambda_{\mathrm{succ}}z_k^{\mathrm{succ}},
\label{eq:local_reward}
\end{equation}
where $\sum_{j\in\mathcal M_i}\eta_j=B_s/n_s$. Here, $z^{\mathrm{mile}}_{j,k}$ indicates the first valid completion of milestone $j$ in action chunk $k$. Milestone rewards replace the target subtask reward to avoid double counting. Subtask checks remain active for progress tracking. Tasks without a selected window or verified milestones retain the original reward.

\noindent\textbf{History Buffer.} After Local RL, we record the diagnosed bottlenecks, selected practice windows, interaction costs, and changes in normal-start full-task success in the History Buffer $\mathcal H$. In subsequent rounds, the Scheduler combines these records with new diagnoses to retain, revise, or skip practice windows, keeping the allocation of training interaction aligned with the current policy's remaining difficulties.

\begingroup
\algrenewcommand\algorithmiccomment[1]{\hfill{\footnotesize\color{gray}$\triangleright$~#1}}
\setlength{\intextsep}{6pt}
\begin{algorithm}[H]
\small
\caption{ReFocus}
\label{alg:repractice}
\begin{algorithmic}[1]
\Require Policy $\pi_\theta$, critic $V_\phi$, tasks $\mathcal T$, Scheduler $\mathcal S$, rounds $R$
\Statex \textbf{Parameters:} phase updates $U_{\mathrm G},U_{\mathrm L}$, rollout probabilities $p_m$
\State $\{G_i\}\gets\mathcal S(\mathcal T)$ \Comment{Task DAGs}
\State $\mathcal H\gets\varnothing$ \Comment{History Buffer}
\For{$r=1,\ldots,R$}
    \State $\mathcal D,\mathcal E\gets\varnothing$ \Comment{Validation and Training Entries}
    \For{$u=1,\ldots,U_{\mathrm G}$} \Comment{\textcolor{red}{\textbf{Global RL}}}
        \State $\tau^{\mathrm G}\gets\operatorname{Rollout}(\pi_\theta)$
        \State $\mathcal E\gets\mathcal E\cup\operatorname{Entries}(\tau^{\mathrm G})$ \Comment{Collect Training Entries}
        \State $\boldsymbol{r}^{\mathrm G}\gets(r_k^{\mathrm G})_{k\in\tau^{\mathrm G}}$ \Comment{Eq.~\eqref{eq:repractice_reward}}
        \State $(\theta,\phi)\gets\operatorname{PPO}(\theta,\phi;\tau^{\mathrm G},\boldsymbol{r}^{\mathrm G})$ \Comment{Global RL Update}
        \State $\mathcal D\gets\mathcal D\cup\operatorname{Eval}(\pi_\theta)$ \textbf{if} $u\bmod 5=0$
    \EndFor
    \State $\{(w_i,\mathcal B_i,\mathcal M_i)\}\gets\mathcal S(\mathcal D,\mathcal H;\mathcal E)$ \Comment{Select Windows}
    \State $\boldsymbol{c}\gets\mathbf 0,\quad\boldsymbol{s}_{\mathrm{pre}}\gets\operatorname{SR}_{\mathrm{task,last}}(\mathcal D)$
    \For{$u=1,\ldots,U_{\mathrm L}$} \Comment{\textcolor{red}{\textbf{Local RL}}}
        \State $m\sim p_m,\tau^{\mathrm L}\gets\operatorname{Rollout}(\pi_\theta;m,\{w_i,\mathcal B_i\})$
        \State $\boldsymbol{r}^{\mathrm L}\gets\operatorname{Reward}(\tau^{\mathrm L};m,\{w_i,\mathcal M_i\})$ \Comment{Eq.~\eqref{eq:local_reward}}
        \State $(\theta,\phi)\gets\operatorname{PPO}(\theta,\phi;\tau^{\mathrm L},\boldsymbol{r}^{\mathrm L})$ \Comment{Local RL Update}
        \State $\boldsymbol{c}\gets\boldsymbol{c}+\operatorname{Cost}(\tau^{\mathrm L})$
        \State $\operatorname{Log}(\operatorname{Eval}(\pi_\theta))$ \textbf{if} $u\bmod 5=0$
    \EndFor
    \State $\mathcal D_{\mathrm{post}}\gets\operatorname{Eval}(\pi_\theta)$ \Comment{Full-Task Evaluation}
    \State $\Delta\boldsymbol{s}\gets\operatorname{SR}_{\mathrm{task}}(\mathcal D_{\mathrm{post}})-\boldsymbol{s}_{\mathrm{pre}}$
    \State $\mathcal H\gets\operatorname{Update}(\mathcal H;r,\mathcal D,\mathcal D_{\mathrm{post}},\{(w_i,\mathcal M_i)\},\boldsymbol{c},\Delta\boldsymbol{s})$
\EndFor
\State \Return $\pi_\theta$
\end{algorithmic}
\end{algorithm}
\endgroup

\vspace{-15pt}
\subsection{Alternating Optimization}
\label{sec:global_local}

We alternate Global RL and Local RL for several rounds to optimize the shared actor $\pi_\theta$ and critic $V_\phi$. Each round comprises full-task training, bottleneck diagnosis and window selection, and local practice. After Local RL, we evaluate the updated policy from normal task starts to assess task progress and task-wise success rates. These results are stored in $\mathcal H$ alongside the selected windows and interaction costs. Together with fresh diagnoses from the next Global RL phase, this history guides whether to retain, revise, or skip practice windows. Repeated assessment keeps subsequent practice aligned with the shared policy's remaining difficulties. Algorithm~\ref{alg:repractice} summarizes this procedure. In Algorithm~\ref{alg:repractice}, $\boldsymbol{r}^{\mathrm G}$ and $\boldsymbol{r}^{\mathrm L}$ collect the scalar rewards of their respective rollouts. The vectors $\boldsymbol{c}$, $\boldsymbol{s}_{\mathrm{pre}}$, and $\Delta\boldsymbol{s}$ collect task-wise practice costs, pre-practice success rates, and post-practice success-rate changes, respectively.

Both phases optimize the policy using PPO~\cite{schulman2017ppo}, while differing in their rollout distributions and rewards. For each trajectory, we compute the GAE estimate $\hat A_k$ using corresponding rewards and normalize valid estimates to obtain $\widetilde A_k$. The actor objective is: 
\begin{equation}
\begin{aligned}
\mathcal J(\theta) &= \mathbb E_{\tau\sim\mathcal P,k}
\Bigl[ \min\Bigl( \rho_k(\theta)\widetilde A_k,\\ &\operatorname{clip}\bigl( \rho_k(\theta),1-\epsilon,1+\epsilon \bigr)\widetilde A_k \Bigr) \Bigr],
\end{aligned}
\label{eq:repractice_ppo}
\end{equation}
where $\mathcal P$ denotes the trajectory distribution induced by the current phase's rollout procedure. The ratio $\rho_k(\theta)$ compares the updated and behavior policies, and $\epsilon$ controls the clipping range. Appendix~\ref{sec:supp_advantage_ppo} derives the Flow-SDE transition ratio and details how the phase rewards enter GAE.

\section{Experiments}
\label{sec:experiments}

\subsection{Experimental Setup}
\label{sec:exp_setup}

\noindent\textbf{Benchmarks and Evaluation Settings.}
We conduct experiments on RoboCasa365 Composite-seen~\cite{robocasa365} and LIBERO-Long~\cite{liu2023libero}. Compared with tasks centered on individual skills, these benchmarks require sustained progress across multiple subgoals, making them suitable for studying intermediate feedback and targeted practice. LIBERO-Long contains 10 tabletop tasks involving multiple objects and goal conditions. We also use 16 Composite-Seen tasks from RoboCasa365 that extend evaluation to diverse kitchen activities involving object transport and appliance interaction. For each benchmark, we initialize $\pi_{0.5}$ through SFT on the official demonstrations and subsequently train a shared policy across all selected tasks. Validation rollouts are used to diagnose bottlenecks and guide practice-window selection. We evaluate each policy on 50 and 100 initial scenarios per task for LIBERO-Long and RoboCasa365. 

\noindent\textbf{Baselines.}
We compare ReFocus with the SFT-initialized $\pi_{0.5}$
policy and four representative methods:
Sparse-RL~\cite{zang2026rlinfvla}, SimpleVLA-RL~\cite{li2026simplevla}, PolicyTrim~\cite{wang2026policytrim}, and StructRL~\cite{yin2026structrl}. Sparse-RL PPO uses a binary success reward for PPO, while StructRL provides a subtask-level progress-reward baseline. All RL baselines use the same training configuration. More details about training and evaluation configurations are provided in Appendix~\ref{sec:supp_seed_variance}. 

\begin{table*}[!t]
\centering
\caption{ 
\textbf{Comparative experiments on RoboCasa365 Composite-Seen~\cite{robocasa365} and LIBERO-Long~\cite{liu2023libero}.} We consistently use $\pi {0.5}$ as base model. Success rates
(\%) are grouped by task horizon. \textbf{Overall} is the mean task success rate. \textbf{$\Delta$} denotes the gain over the best baseline.
}
\vspace{-1mm}
\label{tab:main_results}
\setlength{\tabcolsep}{8pt}
\renewcommand{\arraystretch}{1}
\resizebox{\textwidth}{!}{
\begin{tabular}{l | cccc | cccc}
\toprule
\textbf{Benchmark}
 & \multicolumn{4}{c|}{\textbf{RoboCasa365}}
 & \multicolumn{4}{c}{\textbf{LIBERO-Long}} \\
\cmidrule(lr){1-9}
Horizon Steps
 & 800--1000 & 1000--1400 & 1400--2900 & Overall
 & 250--340 & 340--400 & 400--550 & Overall \\
\midrule

$\pi {0.5}$~\cite{black2025pi05}   & 41.0 & 46.4& 32.9 & 39.3 &94.0 &96.0 &74.0 &90.6 \\
\midrule
Sparse-RL~\cite{zang2026rlinfvla}      & 43.0 & 48.4 & 35.8 & 41.1 & 97.6 & 94.7 & 77.0 & 92.6 \\
SimpleVLA-RL~\cite{li2026simplevla} & 50.7 & 46.8 & 35.6 & 41.9 & 97.2 & 96.7 & 82.0 & 94.0 \\
PolicyTrim~\cite{wang2026policytrim} & 48.7 & 46.0 & 33.5 & 40.3 & 95.6 & 90.7 & 83.0 & 91.6 \\
StructRL~\cite{yin2026structrl}    & 54.0 & 51.2 & 39.4 & 45.8 & 98.0 & 98.0 & 89.0 & 96.2 \\
\rowcolor{lighterrow}
ReFocus& 59.7 & 54.2 & 46.4 & 51.3 & 99.2 & 98.7 & 93.0 & 97.8 \\
\rowcolor{oursrow}
\quad $\Delta$  & \textcolor{ForestGreen}{+5.7} & \textcolor{ForestGreen}{+3.0} & \textcolor{ForestGreen}{+7.0} & \textcolor{ForestGreen}{\textbf{+5.5}} & \textcolor{ForestGreen}{+1.2} & \textcolor{ForestGreen}{+0.7} & \textcolor{ForestGreen}{+4.0} & \textcolor{ForestGreen}{\textbf{+1.6}} \\
\bottomrule
\end{tabular}
}
\vspace{-10pt}
\end{table*}

\subsection{Main Results}
\label{sec:main_results}

\noindent\textbf{Overall Performance.}
Table~\ref{tab:main_results} compares ReFocus with SFT and representative RL post-training methods. ReFocus achieves the highest overall success rates on both benchmarks, reaching 51.3\% on RoboCasa365 and 97.8\% on LIBERO-Long, exceeding the SFT baseline by 12.0 and 7.2 points, respectively. Compared with Sparse-reward PPO~\cite{zang2026rlinfvla}, it improves by 10.2 and 5.2 points. ReFocus also outperforms StructRL~\cite{yin2026structrl} by 5.5 and 1.6 points, respectively. The gains over both sparse- and structured-reward baselines support the effectiveness of combining progress feedback with adaptive practice.

\noindent\textbf{Performance across Task Horizons.}
ReFocus achieves the highest success rate in all six horizon groups. The largest gains over StructRL occur in the longest-horizon group of each benchmark: success increases from 39.4\% to 46.4\% on RoboCasa365 tasks with horizons of 1,400--2,900 steps, and from 89.0\% to 93.0\% on LIBERO-Long tasks with horizons of 400--550 steps. The larger gains on longer tasks suggest that progress-guided adaptive practice is particularly beneficial for sustained multi-stage execution, while improvements in the shorter-horizon groups demonstrate its applicability across different task lengths.

\begin{table}[t]
\centering
\footnotesize
\setlength{\tabcolsep}{0pt}
\renewcommand{\arraystretch}{1}
\caption{\textbf{Ablation of Progress Reward and Practice Window on LIBERO-Long.} Checkmarks indicate enabled components. Success rates (\%) are grouped by task horizon.}
\label{tab:ablation_reward}
\begin{tabular}{@{}>{\centering\arraybackslash}p{0.25\columnwidth}>{\centering\arraybackslash}p{0.25\columnwidth}*{4}{>{\centering\arraybackslash}p{0.125\columnwidth}}@{}}
\toprule
\multicolumn{2}{c}{Components} & \multicolumn{4}{c}{Horizon Steps} \\
\cmidrule(lr){1-2}\cmidrule(lr){3-6}
Progress Reward & Practice Window & 250--340 & 340--400 & 400--550 & Overall \\
\midrule
-- & -- & 97.6 & 94.7 & 77.0 & 92.6 \\
$\checkmark$ & -- & 98.0 & 98.7 & 91.0 & 96.8\\
-- & $\checkmark$ & 98.0 & 95.3 & 79.0 & 93.4 \\
\rowcolor{oursrow}
$\checkmark$ & $\checkmark$ & \textbf{99.2} & \textbf{98.7} & \textbf{93.0} & \textbf{97.8} \\
\bottomrule
\end{tabular}
\vspace{-10pt}
\end{table}

\subsection{Ablation Study}
\label{sec:ablations}

The main benchmark results demonstrate the effectiveness of ReFocus. We further examine Progress Reward, Practice Window, Local RL rollout mixtures, and the allocation of progress feedback. We also examine credit assignment and performance gains across optimization rounds.

\begin{table}[t]
\centering
\footnotesize
\setlength{\tabcolsep}{0pt}
\renewcommand{\arraystretch}{1.0}
\caption{\textbf{Ablation of Rollout Modes during Local RL on
LIBERO-Long.} Checkmarks indicate enabled modes. Success rates
(\%) are grouped by task horizon.}
\label{tab:ablation_mixture}

\begin{tabularx}{\columnwidth}{
    @{}*{7}{>{\centering\arraybackslash}X}@{}
}
\toprule
\multicolumn{3}{c}{Rollout modes}
& \multicolumn{4}{c}{Horizon Steps } \\
\cmidrule(lr){1-3}\cmidrule(lr){4-7}
Normal & Suffix & Prefix
& 250--340 & 340--400 & 400--550 & Overall \\
\midrule
$\checkmark$ & -- & --
& 98.0 & 96.0 & 88.0 & 95.4 \\

$\checkmark$ & $\checkmark$ & --
& 98.0 & 96.7 & 92.0 & 96.4 \\

$\checkmark$ & -- & $\checkmark$
& 98.8 & 98.0 & 90.0 & 96.8 \\

\rowcolor{oursrow}
{\boldmath$\checkmark$} &
{\boldmath$\checkmark$} &
{\boldmath$\checkmark$} &
\textbf{99.2} & \textbf{98.7} &
\textbf{93.0} & \textbf{97.8} \\
\bottomrule
\end{tabularx}
\vspace{-10pt}
\end{table}

\noindent\textbf{Component Ablation.} Table~\ref{tab:ablation_reward} examines Progress Reward and Practice Window on LIBERO-Long. When Progress Reward is disabled, subtask checks and progress records remain available for diagnosis and window selection. From the Sparse-RL PPO baseline of 92.6\%, adding Progress Reward or Practice Window alone increases success to 96.8\% and 93.4\%, respectively. Combining both reaches 97.8\%. On the longest tasks, the combination achieves 93.0\%, compared with 91.0\% and 79.0\% when using Progress Reward or Practice Window alone. These results suggest that stage-balanced progress feedback and bottleneck practice play complementary roles in improving long-horizon execution.

\noindent\textbf{Rollout Modes During Local RL.} Table~\ref{tab:ablation_mixture} compares different Local RL rollout mixture modes while retaining Global RL. In row order, the sampling probabilities for (\emph{Normal}, \emph{Suffix}, \emph{Prefix}) are $(1,0,0)$, $(0.5,0.5,0)$, $(0.5,0,0.5)$, and $(0.5,0.25,0.25)$. Modes are sampled at rollout resets, with normal-start full-task rollouts as fallback for tasks without an eligible window. \emph{Normal window}-only practice achieves 95.4\% success, increasing to 96.4\% with \emph{suffix window}, 96.8\% with \emph{prefix window}, and 97.8\% with both. On the longest tasks, Suffix improves success from 88.0\% to 92.0\%, compared with 90.0\% for Prefix. These results suggest that integrating bottleneck practice with preceding and subsequent task execution improves full-task performance.

\begin{table}[t]
\centering
\footnotesize
\setlength{\tabcolsep}{4pt}
\renewcommand{\arraystretch}{1}
\caption{\textbf{Ablation of reward feedback on LIBERO-Long.} Success rates (\%) are grouped by task horizon.}
\label{tab:ablation_progress_feedback}
\resizebox{\columnwidth}{!}{%
\begin{tabular}{@{}lcccc@{}}
\toprule
\multirow[c]{2}{*}{Reward feedback} & \multicolumn{4}{c}{Horizon Steps} \\
\cmidrule(lr){2-5}
& 250--340 & 340--400 & 400--550 & Overall \\
\midrule
Sparse reward~\cite{zang2026rlinfvla} & 98.0 & 95.3 & 79.0 & 93.4 \\
Structured reward~\cite{yin2026structrl} & 98.4 & 97.3 & 84.0 & 95.2 \\
\rowcolor{oursrow}
Progress reward (ours) & \textbf{99.2} & \textbf{98.7} & \textbf{93.0} & 
\textbf{97.8} \\
\bottomrule
\end{tabular}%
}
\vspace{-10pt}
\end{table}

\begin{figure*}[!t]
    \centering
    \includegraphics[width=\linewidth]{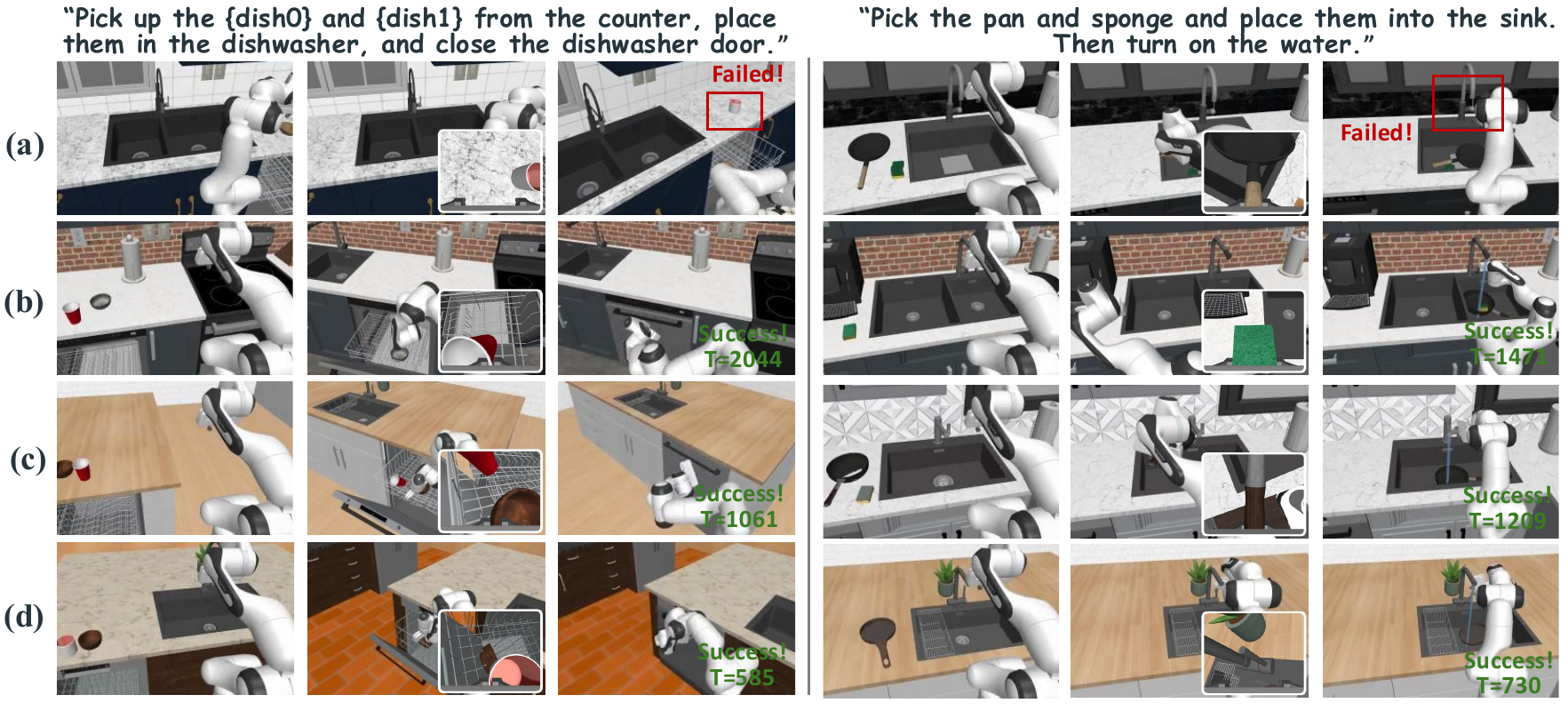}
    \caption{\textbf{Qualitative results on RoboCasa365.} Rows (a)--(d) show rollouts from the SFT policy and policies after Alternating Optimization rounds 1--3, respectively.  Red boxes highlight failures. T denotes the first successful environment step.}
    \label{fig:robocasa_qualitative}
    \vspace{-10pt}
\end{figure*}

\noindent\textbf{Effect of Stage-Balanced Rewards.} Table~\ref{tab:ablation_progress_feedback} compares reward feedback in Global RL while keeping the Local RL configuration unchanged. Beyond guiding Global RL, subtask-level progress records help the Scheduler identify bottlenecks and select practice windows.  Different reward feedback can also lead to different policy rollouts and subsequent window selections. The three settings use binary terminal feedback~\cite{zang2026rlinfvla}, uniform per-subtask rewards~\cite{yin2026structrl}, and the proposed stage-balanced Progress Reward. Overall success increases from 93.4\% with binary terminal feedback to 95.2\% with uniform per-subtask rewards and 97.8\% with stage-balanced rewards. For the multi-stage task \emph{e.g.,} \texttt{MugInMicrowave} , the three settings succeed in 34, 38, and 46 out of 50 trials, respectively.

\begin{figure}[t]
    \centering
    \includegraphics[width=\columnwidth]{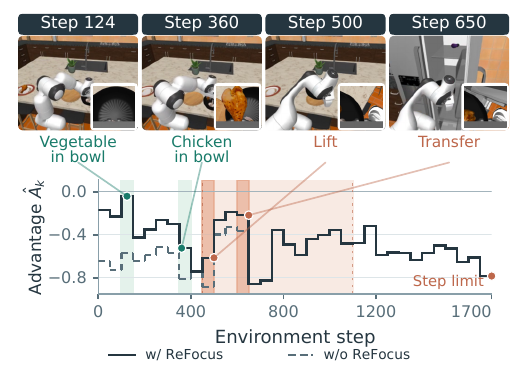}
    \caption{\textbf{Credit assignment with Progress Reward.} With fixed critic predictions, Progress Reward increases the GAE advantages of completed subtasks and milestones in a failed rollout.}
    \label{fig:robocasa_progress_reward_gae}
    \vspace{-10pt}
\end{figure}

\noindent\textbf{Learning from Failed Rollouts.} Figure~\ref{fig:robocasa_progress_reward_gae} visualizes GAE advantages on a failed rollout under sparse binary rewards and Progress Reward, using the same trajectory and critic model. Although the policy completes both food-placement subtasks, it fails to finish the full task and therefore receives no task-success reward under sparse feedback. Progress Reward instead credits these intermediate completions, increasing the advantages of the corresponding action chunks and their preceding actions. The resulting advantage shifts illustrate how progress and milestone rewards provide credit for partial progress even when the full task fails.

\begin{figure}[t]
    \centering
    \includegraphics[width=\columnwidth]{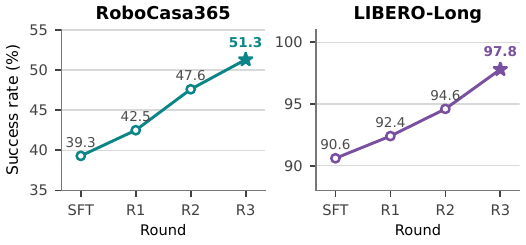}
    \caption{\textbf{Full-task success rate across Alternating Optimization rounds.} Success rates improve from the SFT baseline over three rounds (R1--R3) of alternating Global RL and Local RL.}
    \label{fig:alternating_rounds}
    \vspace{-15pt}
\end{figure}

\begin{figure*}[t]
    \centering
    \includegraphics[width=\textwidth]{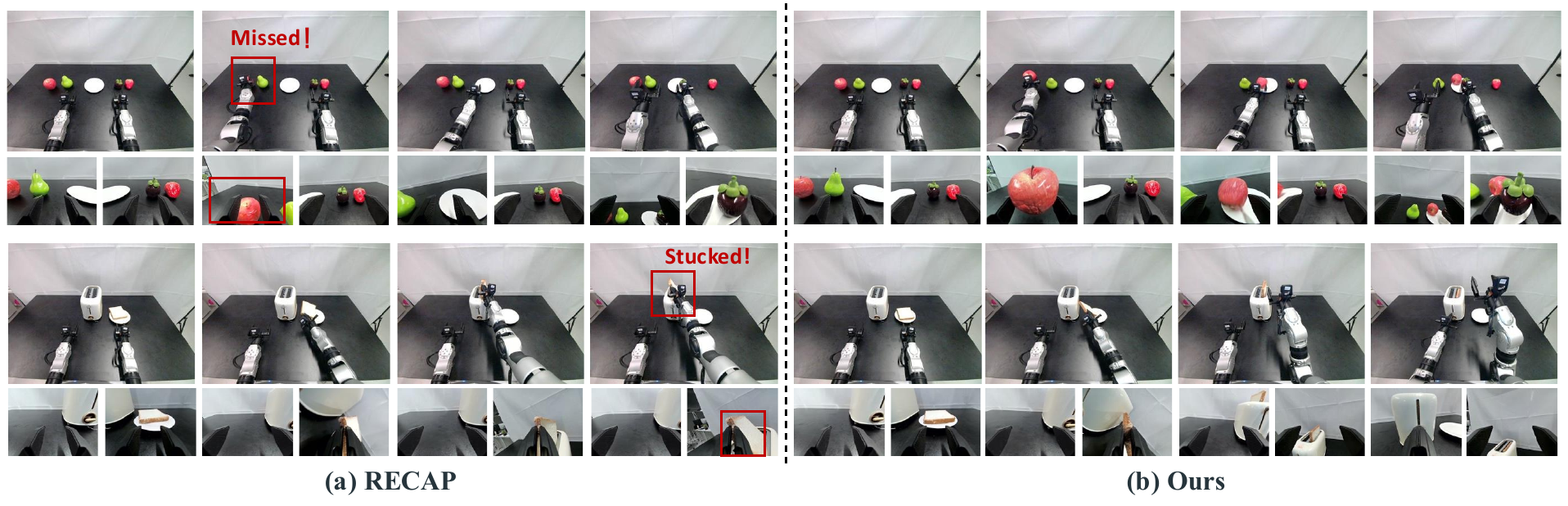}
    \caption{\textbf{Qualitative comparison on real-world tasks.} RECAP misses a grasp in \texttt{PickFruit} (top) and gets stuck in \texttt{BreadBaker} (bottom). ReFocus-based adaptation completes both tasks. Smaller images show wrist-camera views. Red boxes mark failures.}
    \label{fig:real_robot_comparison}
    \vspace{-10pt}
\end{figure*}

\noindent\textbf{Improvements across Alternating Optimization Rounds.} Figure~\ref{fig:robocasa_qualitative} illustrates the effect of successive RL rounds on dishwasher loading and pan pre-soaking. The SFT baseline fails to grasp the cup in \texttt{LoadDishwasher} and to turn on the faucet in  \texttt{PreSoakPan}. After RL training, the policy completes both tasks successfully. Across the illustrated successful rollouts from rounds 1--3, first-success steps decrease from 2044 to 1061 and 585 for \texttt{LoadDishwasher}, and from 1471 to 1209 and 730 for \texttt{PreSoakPan}. The History Buffer allows the Scheduler to account for the costs and outcomes of previous practice when adjusting subsequent practice allocation, helping the policy learn to connect subtasks smoothly. Figure~\ref{fig:alternating_rounds} further shows increasing full-task success across rounds. Success rate on LIBERO-Long rises from 90.6\% at SFT initialization to 92.4\%, 94.6\%, and 97.8\%, while RoboCasa365 improves from 39.3\% to 42.5\%, 47.6\%, and 51.3\%.

\subsection{Real-World Experiments}
\label{sec:real_robot}

\begin{figure}[t]
    \centering
    \includegraphics[width=\linewidth]{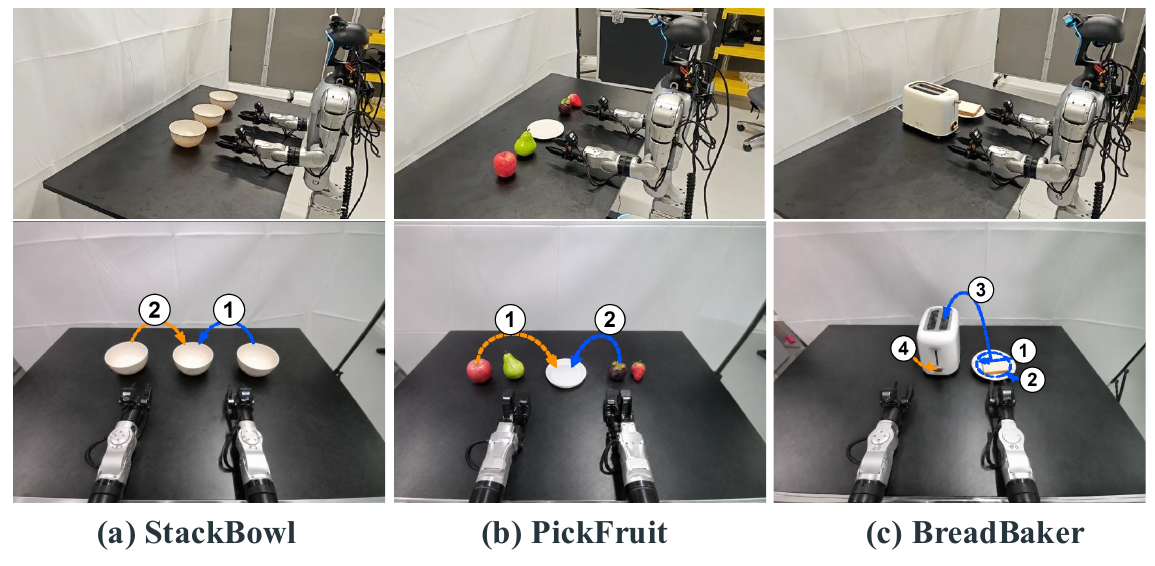}
    \caption{\textbf{Real-world task setups.} We conduct three long-horizon tasks, with circled numbers indicating the order of execution.}
    \label{fig:real_robot_tasks}
    \vspace{-10pt}
\end{figure}

\noindent\textbf{Experimental Setup.} We deploy the policies on a Unitree G1-D humanoid robot. The policy takes RGB images from the right head-mounted camera and both wrist cameras, and outputs joint positions for both arms and gripper actions. We conduct experiments on three tasks, as shown in Fig.~\ref{fig:real_robot_tasks}. For each task, we collect 50 human demonstrations to initialize the $\pi_{0.5}$ policy, followed by 50 autonomous rollouts.

\noindent\textbf{Baselines.} We compare SFT, RECAP~\cite{physicalintelligence2025recap}, and ReFocus-based adaptation. Since online PPO rollouts and intermediate-state resets are costly on physical robots, we adopt RECAP for post-training. GPT-6 Astra~\cite{openai2026gpt6astra} serves as the Scheduler, identifying subtask progress and bottlenecks from rollout observations. We incorporate Progress Rewards when training the value function and computing advantages. We also upsample bottleneck segments alongside full-task rollouts. Each policy is evaluated over three runs of 20 trials per task.

\noindent\textbf{Quantitative Results.} Table~\ref{tab:real_world_results} reports average success rates per task. ReFocus achieves 100\%, 100\%, and 95\% success on \texttt{StackBowl}, \texttt{PickFruit}, and \texttt{BreadBaker}, respectively. The largest gain occurs on \texttt{BreadBaker}, which requires coordinating bread grasping, transport, insertion, and switch operation. This result suggests that combining progress feedback with targeted practice can help the policy overcome bottlenecks and complete long-horizon tasks more reliably.

\begin{table}[t]
\centering
\caption{\textbf{Real-world evaluation results.} We present average success rates over three individual runs.}
\label{tab:real_world_results}
\setlength{\tabcolsep}{4pt}
\renewcommand{\arraystretch}{1.1}
\resizebox{\columnwidth}{!}{%
\begin{tabular}{lccc}
\toprule
\textbf{Method} & \texttt{StackBowl} & \texttt{PickFruit} & \texttt{BreadBaker} \\
\midrule
$\pi_{0.5}$~\cite{black2025pi05} & 16/20 (80\%) & 15/20 (75\%) & 6/20 (30\%) \\
RECAP~\cite{physicalintelligence2025recap} & 18/20 (90\%) & 17/20 (85\%) & 14/20 (70\%) \\
ReFocus & \textbf{20/20 (100\%)} & \textbf{20/20 (100\%)} & \textbf{19/20 (95\%)} \\
\bottomrule
\end{tabular}}
\vspace{-10pt}
\end{table}

\noindent\textbf{Qualitative Results.} Figure~\ref{fig:real_robot_comparison} shows representative rollouts. In \texttt{PickFruit}, RECAP fails to grasp a fruit, whereas ReFocus places both fruits on the plate. In \texttt{BreadBaker}, the Scheduler analyzes rollouts collected using the SFT-initialized $\pi_{0.5}$ policy and identifies bread grasping and insertion as bottlenecks. After further training and deployment, our policy completes the insertion and activates the toaster, while RECAP still stalls during insertion. These examples illustrate how progress-guided feedback and targeted practice can help address difficult subtasks and support successful multi-stage execution. Additional implementation details are provided in Appendix~\ref{sec:real_world_adaptation}.

\section{Conclusion}
\label{sec:conclusion}

In this work, we present ReFocus, a progress-guided multi-round RL framework for long-horizon manipulation that alternates Global and Local RL to improve a shared VLA policy. Specifically, during Global RL, a DAG-based Progress Reward balances reward allocation across stages for policy training and supports bottleneck diagnosis. After identifying bottlenecks, Adaptive Practice Windows concentrate additional interaction on the corresponding bottlenecks during Local RL. Experiments on LIBERO-Long and RoboCasa365 show that ReFocus improves the SFT baseline from 90.6\% and 39.3\% to 97.8\% and 51.3\%, respectively. A RECAP-based adaptation extends the framework to three real-world tasks, achieving 95\% success on \textsc{BreadBaker} compared with 30\% for the SFT-initialized $\pi_{0.5}$ policy.

\section*{LLM Usage Statement}
OpenAI Codex assisted with literature discovery, citation checking and language editing of the manuscript. The Practice Scheduler uses Frontier LLM to analyze development rollouts and propose practice windows, as described in the method and supplementary material. The authors are responsible for the final manuscript, implementation, experiments, and scientific claims.

\clearpage
\onecolumn
\appendix

\begin{center}
    {\Large\bfseries\thetitle\par}
    \vspace{0.5em}
    {\large\bfseries Supplementary Material\par}
    \ifboolexpr{not togl {cvprfinal} and not togl {cvprrebuttal}}{%
        \vspace{0.9em}
        {\normalsize Anonymous \confName~submission\par}
    }{}
\end{center}
\vspace{1.4em}

\begingroup
\definecolor{suppcontentsblue}{RGB}{10,24,96}
\hypersetup{linkcolor=suppcontentsblue}
\newcommand{\suppcontentssection}[2]{%
    \par\addvspace{10pt}\noindent
    \makebox[\linewidth][l]{%
        \hyperref[#1]{\bfseries\makebox[2.2em][l]{\ref*{#1}}#2}%
        \hfill\textbf{\pageref*{#1}}}\par}
\newcommand{\suppcontentssubsection}[2]{%
    \par\noindent\hspace*{2.2em}%
    \hyperref[#1]{\makebox[2.8em][l]{\ref*{#1}}#2}%
    \nobreak\leaders\hbox to .7em{\hss.\hss}\hfill
    \nobreak\hspace{.7em}\pageref*{#1}\par}
{\LARGE\bfseries Table of Contents\par}
\vspace{3pt}\hrule height .5pt
\vspace{4pt}
\begin{list}{}{\setlength{\leftmargin}{1.5em}\setlength{\rightmargin}{1.5em}%
    \setlength{\topsep}{0pt}\setlength{\parsep}{0pt}\setlength{\itemsep}{0pt}}
\item[]\fontsize{12}{18}\selectfont
\suppcontentssection{sec:supp_failure_stages}{Failure-Stage Analysis}
\suppcontentssection{sec:supp_algorithm}{Method Details}
\suppcontentssubsection{sec:supp_dag_reward}{Task Specifications and Completion Checks}
\suppcontentssubsection{sec:supp_restore_details}{State Storage and Restoration}
\suppcontentssubsection{sec:supp_prompt_design}{Scheduler Inputs and Prompt Design}
\suppcontentssection{sec:supp_seed_variance}{Experimental Details}
\suppcontentssubsection{sec:supp_implementation}{Training Configurations}
\suppcontentssubsection{sec:supp_data_settings}{Data Sampling and Evaluation}
\suppcontentssubsection{sec:supp_multiseed_training_and_evaluation}{Multi-seed Training and Evaluation}
\suppcontentssubsection{sec:supp_window_history_ablation}{Ablation Study on Window Selection across Rounds}
\suppcontentssubsection{sec:supp_scheduler_model_comparison}{Ablation Study on the Scheduler}
\suppcontentssubsection{sec:supp_reward_allocation_control}{Reward Allocation Analysis}
\suppcontentssubsection{sec:supp_practice_allocation}{Practice Allocation Analysis}
\suppcontentssection{sec:real_world_adaptation}{Real-World Experiments}
\suppcontentssubsection{sec:supp_real_tasks}{Real-World Task Procedures}
\suppcontentssubsection{sec:supp_recap_adaptation}{RECAP-Based Adaptation}
\suppcontentssubsection{sec:supp_real_value}{Value Model Visualization}
\suppcontentssection{sec:supp_qualitative}{Additional Qualitative Results}
\suppcontentssection{sec:supp_limitations}{Limitations and Future Work}
\end{list}
\vspace{10pt}\hrule height .5pt
\endgroup
\clearpage

\iftoggle{cvprfinal}{}{\linenumbers}

\raggedbottom

\section{Failure-Stage Analysis}
\label{sec:supp_failure_stages}

Fig.~\ref{fig:robocasa_failure_stages} summarizes task-specific failure patterns during Global RL. We collect five normal-start validation rollouts per task at each of five checkpoints across the 16 RoboCasa365 Composite-Seen tasks, yielding 400 rollouts and 228 failures. Numbers inside the bars indicate failure counts, and the right-hand labels show failed/total rollouts per task. Figure~\ref{fig:robocasa_task_bottlenecks} highlights three tasks with failures concentrated at different stages. In \texttt{GetToastedBread}, 17 of 19 failures occur at the initial toaster-activation stage. In \texttt{KettleBoiling}, 9 of 11 failures occur at burner activation after the kettle has been placed on the stove. In \texttt{StirVegetables}, 10 of 15 failures occur at the final stirring stage. These results reveal distinct bottlenecks at early, intermediate, and late stages across tasks. Despite these differences in completed subtasks and remaining difficulties, all failed rollouts receive the same terminal failure feedback. This feedback alone provides little guidance on which actions made valid progress or which stages require further practice. These observations motivate using subtask-level progress to identify task-specific bottlenecks and direct additional practice toward the stages that remain difficult.

\newlength{\suppfailurepanelheight}
\begin{figure}[!htb]
\setlength{\suppfailurepanelheight}{0.525\linewidth}
\centering
\begin{minipage}[t]{0.48\linewidth}
\vspace{0pt}
\centering
\includegraphics[height=\suppfailurepanelheight,keepaspectratio]{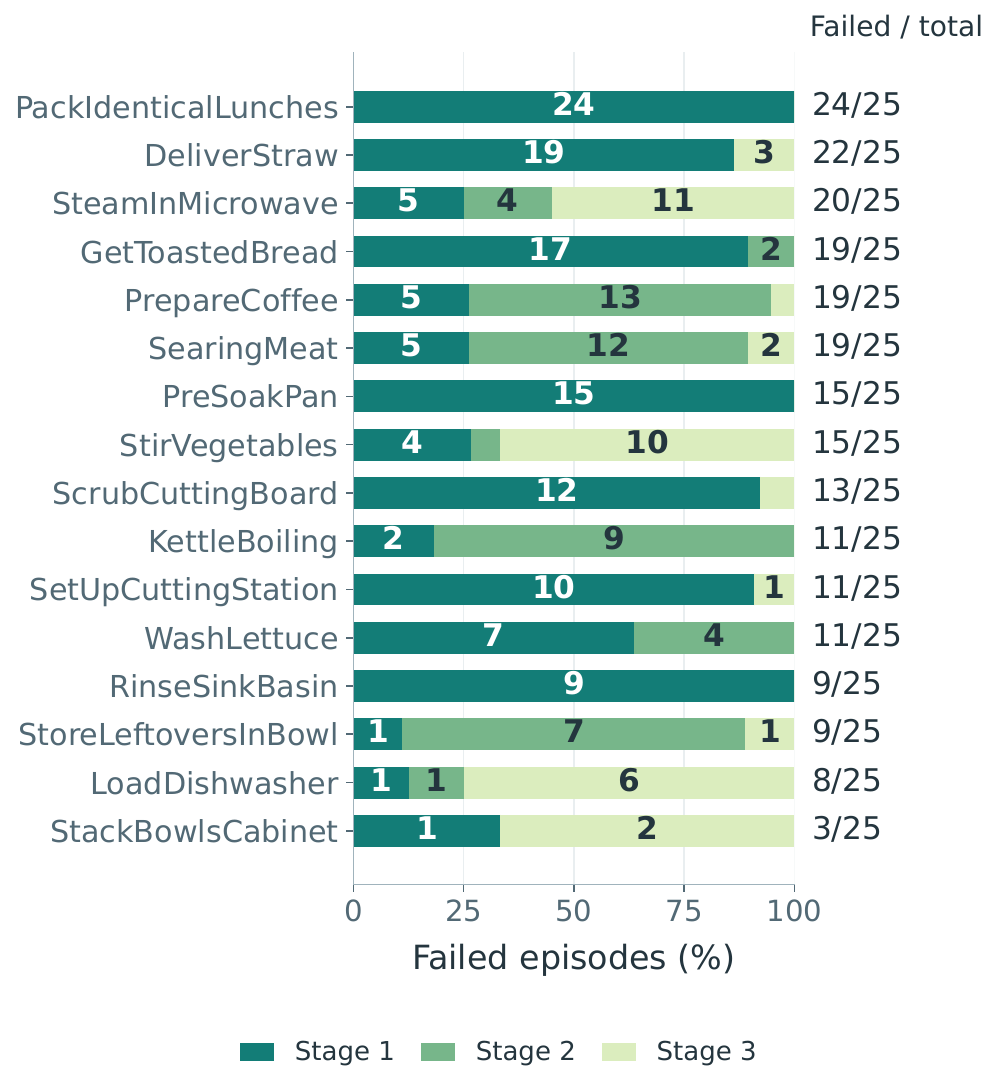}
\caption{\textbf{Failure-stage distributions on RoboCasa365.} Bars show the first unfinished stage in archived validation rollouts. }
\label{fig:robocasa_failure_stages}
\end{minipage}\hfill
\begin{minipage}[t]{0.48\linewidth}
\vspace{0pt}
\centering
\includegraphics[height=\suppfailurepanelheight,keepaspectratio]{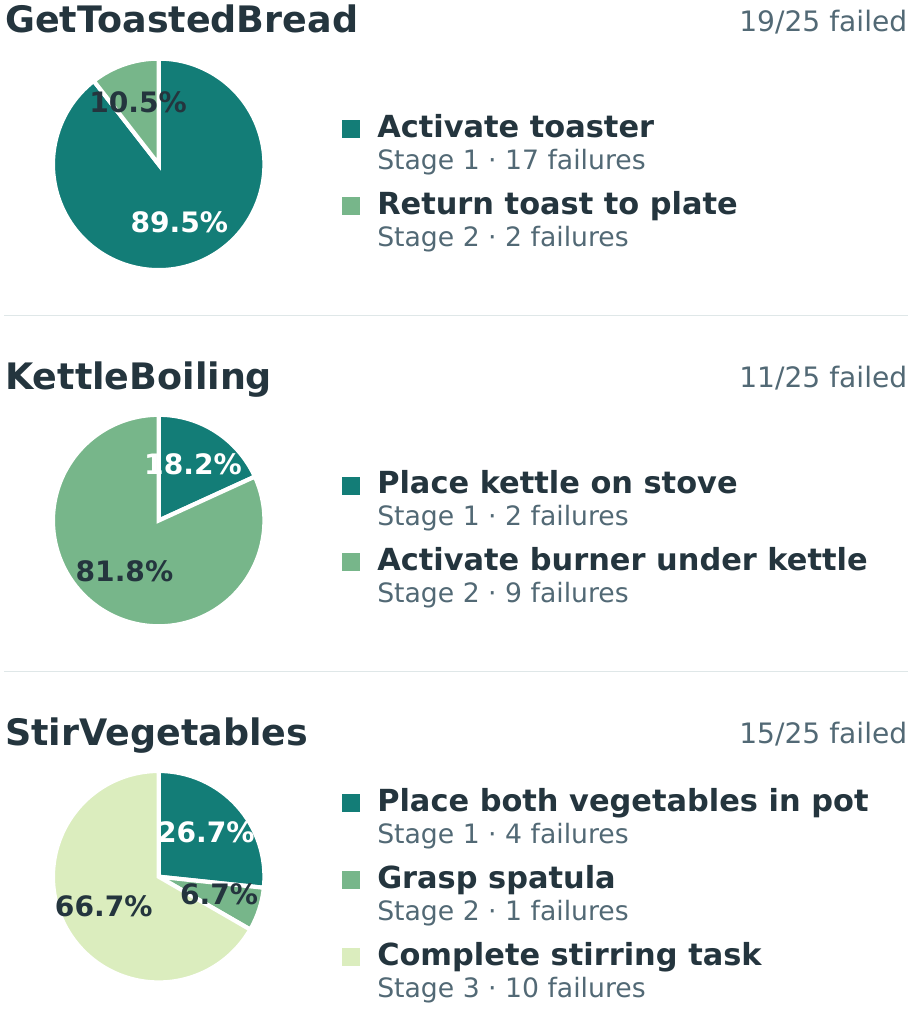}
\caption{\textbf{Task-specific bottlenecks in three representative tasks.} Stage descriptions and counts are provided alongside.}
\label{fig:robocasa_task_bottlenecks}
\label{fig:robocasa_typical_failure_pies}
\end{minipage}
\end{figure}

\section{Method Details}
\label{sec:supp_algorithm}

In this section, we provide more details about the task specifications and completion checks (\textbf{\S}\ref{sec:supp_dag_reward}), the state-storage and restoration procedure (\textbf{\S}\ref{sec:supp_restore_details}), and the Scheduler inputs and complete prompt templates (\textbf{\S}\ref{sec:supp_prompt_design}).

\subsection{Task Specifications and Completion Checks}
\label{sec:supp_dag_reward}

We implement the completion conditions proposed by the Scheduler using task-specific simulator checks, including object--receptacle relations, fixture states, and cumulative activity checks. Tables~\ref{tab:dag_robocasa} and~\ref{tab:dag_libero} list the stage definitions for RoboCasa365 and LIBERO-Long, respectively. At each reward evaluation, completion checks use the preceding progress state to determine eligibility. Newly valid events are collected before advancing the active stage, so future-stage predicates are not credited in the same evaluation. Each event is recorded once and contributes to the reward of its action chunk.

Table~\ref{tab:milestone_window_examples} illustrates milestone refinements within the target subtasks of six recorded practice windows. The examples use available simulator signals to provide finer feedback for object transport, fixture operation, insertion, and continuous activity. Table~\ref{tab:milestone_libero_examples} gives three corresponding examples for LIBERO-Long, with target subtasks taken directly from Table~\ref{tab:dag_libero}.

\begin{table}[t]
\centering
\small
\caption{\textbf{Refined Task DAGs for RoboCasa365 Composite-Seen.} Each task is decomposed into stages of verifiable progress predicates. Predicates within braces belong to the same stage and may be credited in any order. Arrows indicate prerequisites between stages. For tasks with duration requirements, completion is determined by the \texttt{task\_complete} signal.}
\label{tab:dag_robocasa}
\begin{tabular}{>{\raggedright\arraybackslash}p{0.2\textwidth}>{\raggedright\arraybackslash}p{0.80\textwidth}}
\toprule
Task & Progress predicates by stage \\
\midrule
DeliverStraw & \{\texttt{grasp\_straw}\} $\rightarrow$ \{\texttt{straw\_in\_glass\_cup}\} \\
GetToastedBread & \{\texttt{toaster\_on}\}  $\rightarrow$ \{\texttt{toast\_on\_plate}\} \\
KettleBoiling & \{\texttt{grasp\_kettle}\} $\rightarrow$ \{\texttt{kettle\_on\_stove}\} $\rightarrow$ \{\texttt{kettle\_on\_active\_burner}\} \\
LoadDishwasher & \{\texttt{dish0\_on\_rack}, \texttt{dish1\_on\_rack}\} $\rightarrow$ \{\texttt{door\_half\_closed}\} $\rightarrow$ \{\texttt{dishwasher\_closed}\} \\
PackIdenticalLunches & \{\texttt{vegetable0\_packed}, \texttt{vegetable1\_packed}\} $\rightarrow$ \{\texttt{meat0\_packed}, \texttt{meat1\_packed}\} \\
PreSoakPan & \{\texttt{water\_on}, \texttt{pan\_in\_sink}, \texttt{sponge\_in\_sink}\} \\
PrepareCoffee & \{\texttt{grasp\_mug}\} $\rightarrow$ \{\texttt{mug\_at\_machine}\} $\rightarrow$ \{\texttt{machine\_turned\_on}\} \\
RinseSinkBasin & \{\texttt{water\_on}\} $\rightarrow$ \{\texttt{washed\_left}, \texttt{washed\_center}, \texttt{washed\_right}\} \\
ScrubCuttingBoard & \{\texttt{grasp\_sponge}\} $\rightarrow$ \{\texttt{contact\_5steps}, \texttt{sweep\_range\_0p1m}\} \\
SearingMeat & \{\texttt{meat\_in\_pan}\} $\rightarrow$ \{\texttt{grasp\_pan}\} $\rightarrow$ \{\texttt{pan\_on\_active\_knob}\} \\
SetUpCuttingStation & \{\texttt{meat\_on\_board}, \texttt{knife\_on\_board}\} \\
StackBowlsCabinet & \{\texttt{grasp\_any\_bowl}\} $\rightarrow$ \{\texttt{bowls\_stacked}\} $\rightarrow$ \{\texttt{any\_bowl\_in\_cabinet}\} \\
SteamInMicrowave & \{\texttt{veg\_in\_bowl}\} $\rightarrow$ \{\texttt{bowl\_in\_micro}\} $\rightarrow$ \{\texttt{door\_closed}\} $\rightarrow$ \{\texttt{turn\_on\_microwave}\} \\
StirVegetables & \{\texttt{veg1\_in\_pot}, \texttt{veg2\_in\_pot}\} $\rightarrow$ \{\texttt{spatula\_grasped}\} $\rightarrow$ \{\texttt{stir\_t1}\} $\rightarrow$ \{\texttt{task\_complete}\} \\
StoreLeftoversInBowl & \{\texttt{chicken\_in\_bowl}, \texttt{vegetable\_in\_bowl}\} $\rightarrow$ \{\texttt{grasp\_bowl}\} $\rightarrow$ \{\texttt{bowl\_in\_fridge}\} \\
WashLettuce & \{\texttt{water\_on}\} $\rightarrow$ \{\texttt{lettuce\_under\_water}\} $\rightarrow$ \{\texttt{wash\_t10}\} $\rightarrow$ \{\texttt{task\_complete}\} \\
\bottomrule
\end{tabular}
\par\smallskip
\end{table}

\begin{table}[t]
\centering
\small
\caption{\textbf{Task DAGs for LIBERO-Long.} Each task is decomposed into stages of verifiable progress predicates. Predicates within braces belong to the same stage and may be credited in any order. Arrows indicate prerequisite relations between stages. }
\label{tab:dag_libero}
\begin{tabular}{p{0.2\textwidth}p{0.8\textwidth}}
\toprule
Task & Progress predicates by stage \\
\midrule
HeatMokaPot & \{\texttt{turnon}(\texttt{flat\_stove\_1}), \texttt{on}(\texttt{moka\_pot\_1}, \texttt{flat\_stove\_1\_cook\_region})\} \\
StoreBowlInDrawer & \{\texttt{in}(\texttt{akita\_black\_bowl\_1}, \texttt{white\_cabinet\_1\_bottom\_region})\} \newline $\rightarrow$ \{\texttt{close}(\texttt{white\_cabinet\_1\_bottom\_region})\} \\
MugInMicrowave & \{\texttt{in}(\texttt{white\_yellow\_mug\_1}, \texttt{microwave\_1\_heating\_region})\} $\rightarrow$ \{\texttt{close}(\texttt{microwave\_1})\} \\
MokaPotsOnStove & \{\texttt{on}(\texttt{moka\_pot\_1}, \texttt{flat\_stove\_1\_cook\_region}), \texttt{on}(\texttt{moka\_pot\_2}, \texttt{flat\_stove\_1\_cook\_region})\} \\
SoupCheeseInBasket & \{\texttt{in}(\texttt{alphabet\_soup\_1}, \texttt{basket\_1\_contain\_region}), \newline \texttt{in}(\texttt{cream\_cheese\_1}, \texttt{basket\_1\_contain\_region})\} \\
SoupSauceInBasket & \{\texttt{in}(\texttt{alphabet\_soup\_1},   \texttt{basket\_1\_contain\_region}), \newline \texttt{in}(\texttt{tomato\_sauce\_1}, \texttt{basket\_1\_contain\_region})\} \\
CheeseButterInBasket & \{\texttt{in}(\texttt{cream\_cheese\_1}, \texttt{basket\_1\_contain\_region}), \newline \texttt{in}(\texttt{butter\_1}, \texttt{basket\_1\_contain\_region})\} \\
MugsOnPlates & \{\texttt{on}(\texttt{porcelain\_mug\_1}, \texttt{plate\_1}), \texttt{on}(\texttt{white\_yellow\_mug\_1}, \texttt{plate\_2})\} \\
ArrangeMugAndPudding & \{\texttt{on}(\texttt{porcelain\_mug\_1}, \texttt{plate\_1}), \texttt{on}(\texttt{chocolate\_pudding\_1}, \newline 
\texttt{living\_room\_table\_plate\_right\_region})\} \\
BookInCaddy & \{\texttt{in}(\texttt{black\_book\_1}, \texttt{desk\_caddy\_1\_back\_contain\_region})\} \\
\bottomrule
\end{tabular}
\end{table}

\begin{table}[t]
\centering
\small
\setlength{\tabcolsep}{5pt}
\renewcommand{\arraystretch}{1.1}
\caption{\textbf{Examples of Milestones for Robocasa365 Composite-seen.} Target subtasks are the original predicates from the Task DAGs. Stage indices are zero-based and inclusive. Each target subtask is illustrated with two to four milestones based on available checker signals.}
\label{tab:milestone_window_examples}
\begin{tabular}{@{}p{0.22\textwidth}p{0.24\textwidth}p{0.49\textwidth}@{}}
\toprule
Task and recorded window & Target subtask & Milestones and completion signals \\
\midrule
StoreLeftoversInBowl\newline Stage 1, 600 steps
& \texttt{bowl\_in\_fridge}
& 1. Lift the bowl: grasp status and height increase.\newline
2. Bring it near the fridge: bowl--fridge distance.\newline
3. Place it inside: bowl--fridge containment. \\
\addlinespace[4pt]
KettleBoiling\newline Stage 1, 600 steps
& \texttt{kettle\_on\_active\_burner}
& 1. Reach the corresponding knob: gripper--knob distance.\newline
2. Begin turning it: knob angular displacement.\newline
3. Activate the burner under the kettle: burner state, kettle alignment, and stove contact. \\
\addlinespace[4pt]
StirVegetables\newline Stage $[1,2]$, 700 steps
& \texttt{task\_complete}
& 1. Contact the vegetables: spatula--vegetable contact.\newline
2. Accumulate stirring motion: contact-gated travel of both vegetables.\newline
3. Complete stirring: the original completion predicate. \\
\addlinespace[4pt]
LoadDishwasher\newline Stage 1, 500 steps
& \texttt{dishwasher\_closed}
& 1. Partially close the door: normalized door-joint position.\newline
2. Close the door: the original closure predicate. \\
\addlinespace[4pt]
PrepareCoffee\newline Stage $[0,1]$, 800 steps
& \texttt{machine\_turned\_on}
& 1. Contact the start button: gripper--button contact.\newline
2. Activate the machine: machine-on state. \\
\addlinespace[4pt]
StackBowlsCabinet\newline Stage 1, 500 steps
& \texttt{any\_bowl\_in\_cabinet}
& 1. Lift a bowl: bowl height increase.\newline
2. Bring it near the cabinet: bowl--cabinet distance.\newline
3. Insert it partially: partial containment.\newline
4. Place it inside: full containment. \\
\bottomrule
\end{tabular}
\end{table}

\begin{table}[t]
\centering
\small
\setlength{\tabcolsep}{5pt}
\renewcommand{\arraystretch}{1.1}
\caption{\textbf{Examples of Milestones for LIBERO-Long.} Target subtasks are the original predicates from the Task DAGs. Stage indices are zero-based and inclusive. Each target subtask is illustrated with two to four milestones based on available checker signals.}
\label{tab:milestone_libero_examples}
\begin{tabular}{@{}>{\raggedright\arraybackslash}p{0.18\textwidth}>{\raggedright\arraybackslash}p{0.35\textwidth}>{\raggedright\arraybackslash}p{0.42\textwidth}@{}}
\toprule
Task and recorded window & Target subtask & Milestones and completion signals \\
\midrule
MugInMicrowave \newline Stage 0, 300 steps & \texttt{in(white\_yellow\_mug\_1,}\newline\texttt{microwave\_1\_heating\_region)} & 1. Lift the mug: grasp status and height increase.\newline 2. Align it with the opening: mug pose relative to the heating region.\newline 3. Place it inside: the original mug-containment predicate. \\
\addlinespace[4pt]
StoreBowlInDrawer\newline Stage 1, 250 steps & \texttt{close(}\newline\texttt{white\_cabinet\_1\_bottom\_region)} & 1. Partially close the drawer: drawer-joint position.\newline 2. Close it: the original drawer-closure predicate. \\
\addlinespace[4pt]
BookInCaddy\newline Stage 0, 300 steps & \texttt{in(black\_book\_1,}\newline\texttt{desk\_caddy\_1\_back\_contain\_region)} & 1. Grasp the book: a validated grasp check.\newline 2. Lift it: book height increase while grasped.\newline 3. Align it with the back compartment: book pose relative to the target region.\newline 4. Insert it: the original book-containment predicate. \\
\bottomrule
\end{tabular}
\end{table}

\begin{figure}[t]
\centering
\includegraphics[width=0.8\textwidth]{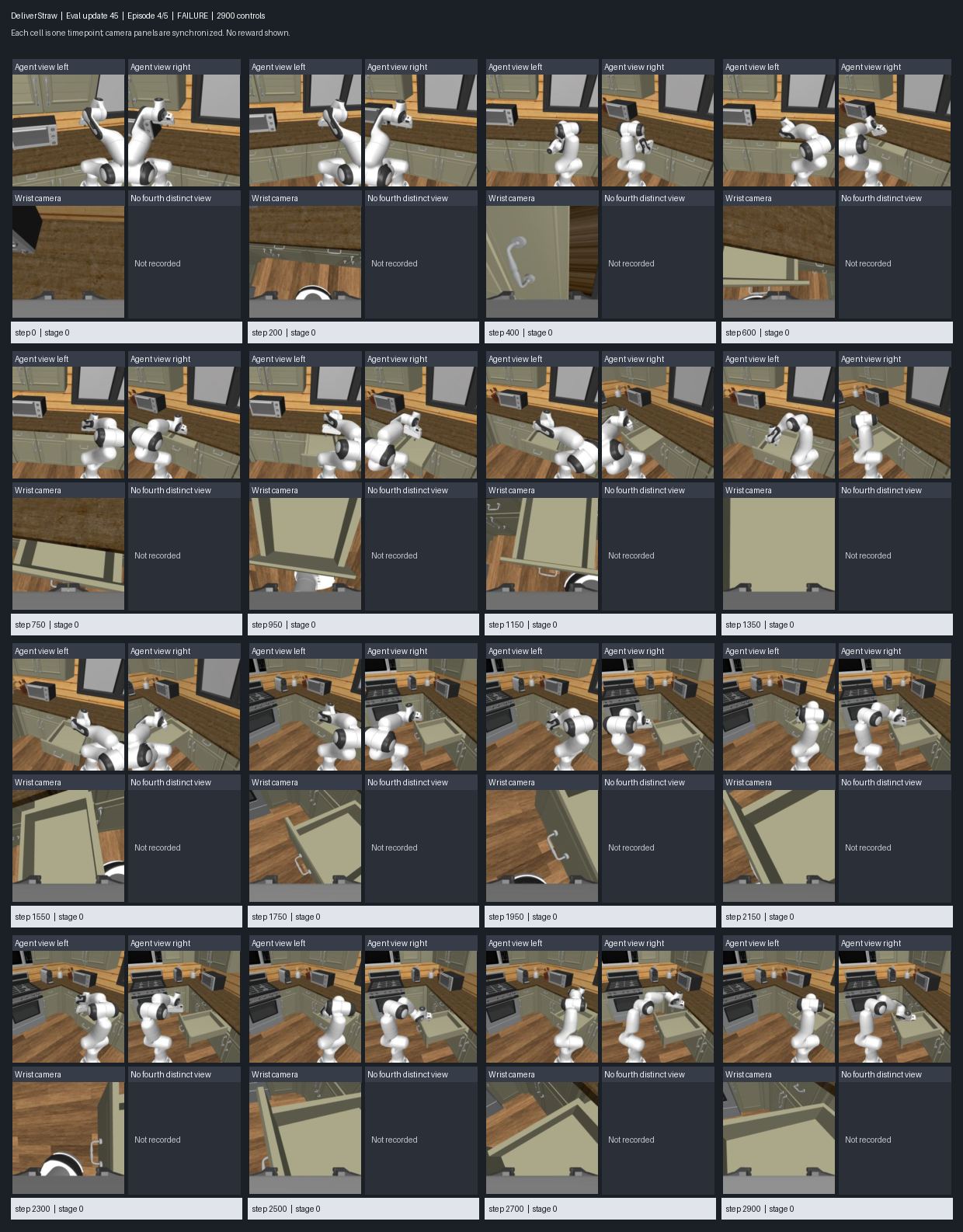}
\caption{\textbf{Visual input for the Scheduler during bottleneck diagnosis.} Each cell shows synchronized left, right, and wrist-camera views, together with the control step and task stage.}
\label{fig:supp_astra_montage}
\end{figure}

\subsection{State Storage and Restoration}
\label{sec:supp_restore_details}
\label{sec:supp_alternating_details}

\noindent \textbf{Training Entry Collection and Storage.} During Global RL, we collect a bounded bank of restorable entries at action-chunk boundaries. For each rollout, we retain the first eligible entry for each unfinished stage encountered. Entries are indexed by round, task, and stage, with up to four records per worker in each bank. Once a practice window is selected, we validate its task identifier and stage bounds, confirm entry availability, and retrieve matching records from the current round. Each entry stores the reset arguments needed to reproduce it: the episode seed, executed action prefix, flattened simulator state, and the reward checker's latched completion flags and timers. Each record also includes the task and round identifiers, source update, stage-definition signature, and scene-XML fingerprint. These fields support entry matching and consistency checks during restoration.

\noindent \textbf{Environment Reset and Entry Restoration.} We directly restore each selected training entry from a snapshot. The snapshot stores the scene XML with runtime object poses, the full MuJoCo integration state, controller state, observation cache, reward-checker state, and environment counters. After reconstructing the scene, we restore these states and verify the entry against its saved record, requiring agreement of the integration state within an absolute tolerance of $10^{-6}$ and zero relative tolerance, as well as equality of the reward checker's completion latches and timers. \footnote{For MuJoCo state-handling interfaces, see \url{https://github.com/google-deepmind/mujoco/blob/main/src/engine/engine_support.c}.} Reaching the window budget is treated as a truncation: the final observation is retained for value bootstrapping, and the GAE recursion stops at the trajectory boundary. 

\subsection{Practice Scheduler: Inputs, History Buffer, and Prompt Templates}
\label{sec:supp_astra_example}
\label{sec:supp_window_details}
\label{sec:supp_history_buffer}
\label{sec:supp_advisor}
\label{sec:supp_prompt_design}

For each task, the Scheduler receives the instruction, ordered stage definitions, progress records, and recorded environment outcomes, together with a $4\times4$ timeline montage of frames sampled along each rollout. Each cell contains synchronized external and wrist-camera views, when available, labeled with the control step and task stage. Figure~\ref{fig:supp_astra_montage} shows an example from \texttt{DeliverStraw}. The diagnosis output is then passed to the Scheduler together with the available training entries.

Previous interventions are supplied through task-level History Buffer records containing the selected window and milestones, realized rollout modes, fresh-interaction and prefix-replay costs, and pre- and post-practice success counts with their evaluation denominators. Records also retain task, round, and checkpoint identifiers. The Scheduler retrieves records from up to two preceding rounds, including evaluations immediately before Local RL, after five Local updates, and at the end of the Local phase. Detailed prompts provided to the Scheduler are presented below.

\begin{promptbox}[unbreakable,title after break={Task DAG Construction Prompt (continued)}]{Task DAG Construction Prompt}
You are a robot task planner. Given the task instruction, scene information, and available completion checks, construct a Task DAG of verifiable subtasks and their dependencies. The DAG defines eligibility for progress rewards, and its completion records support bottleneck diagnosis and practice-window selection.

Requirements:
1. Represent each subtask as a node and each prerequisite relation as a directed edge. The graph must be acyclic.
2. Encode the DAG as ordered stages. Subtasks within a stage must have no dependencies on one another and may be completed in any order. Every subtask in a stage must be credited before the next stage becomes eligible for reward. This specifies credit eligibility, not simultaneous execution.
3. Preserve meaningful dependencies. For example, placing chicken and vegetables in a bowl may share one stage, while placing the loaded bowl on a fridge shelf requires both placements to have been credited.
4. Describe each node as one atomic, physically verifiable event using a short verb phrase, such as grasp an object, place an object in a receptacle, open a fixture, or press a button.
5. Include meaningful intermediate progress. For continuous activities such as stirring, washing, or scrubbing, use verifiable milestones supported by the supplied completion checks. Do not invent simulator signals or thresholds.
6. Exclude redundant retreat motions and a generic task-finished node. The environment checks full-task success separately. Each valid subtask completion is credited once per episode.

Output format requirement: Return only a valid JSON object matching the structure below. Replace the example strings with the proposed subtasks. The outer array lists stages in order, and each inner array lists the subtasks in one stage. Do not include Markdown fences, comments, or text outside the JSON object.
{
  "stages": [
    ["subtask", "another subtask"],
    ["subsequent subtask"]
  ]
}
\end{promptbox}

\begin{promptbox}[unbreakable,title after break={Bottleneck Diagnosis Prompt (continued)}]{Bottleneck Diagnosis Prompt}
You are a visual robotics RL analyst. Given the task instruction, stage definitions, rollout montages, recorded outcomes, and available intervention history, diagnose task progress and recurring bottlenecks during Global RL.

Requirements:
1. Inspect the attached 4x4 timeline montages. Each cell contains a 2x2 camera panel labeled agentview_left, agentview_right, wrist, or Not recorded. Its caption gives the control step and recorded stage.
2. Use the recorded SUCCESS or FAILURE label to determine the episode outcome. Do not override it based on visual appearance or treat an unrecorded view as evidence of failure.
3. Describe visible subtask progress and identify the earliest stage at which failures repeatedly stall. Support the diagnosis with specific observations from the supplied rollouts.
4. Use zero-based stage indices for recurring_stall_stage, consistent with the window-selection response. Return null when no recurring stall can be identified. Do not report reward values.
5. When prior Local interventions are available, compare the Global-end baseline, Local +5, and Local +25 evaluations descriptively. Do not infer causality from these comparisons.
6. State uncertainty explicitly when views, temporal coverage, or other evidence are insufficient for a clear diagnosis.

Output format requirement: Return only a valid JSON object matching the structure below. Replace task_name with the exact input task name. Use a string for visible_progress and uncertainty, an integer or null for recurring_stall_stage, and an array of strings for evidence. The values shown are examples. Do not include Markdown fences, comments, or text outside the JSON object.
{
  "task": "task_name",
  "visible_progress": "observed progress",
  "recurring_stall_stage": 0,
  "evidence": ["supporting observation"],
  "uncertainty": "limitations"
}
\end{promptbox}

\begin{promptbox}[breakable,lines before break=12,title after break={Joint Window-Selection Prompt (continued)}]{Joint Window-Selection Prompt}
You are a practice Scheduler for long-horizon robotic manipulation. Given task diagnoses, available training entries, and prior intervention history, select the next Local RL practice window for each eligible task.

Requirements:
1. Return exactly one window for each eligible task. Do not include ineligible tasks or introduce task names absent from the input.
2. Use zero-based, inclusive start_stage and end_stage indices. Select start_stage only from stages with available training entries. Ensure start_stage <= end_stage and keep both indices within the task's configured stage range.
3. Target the earliest recurring bottleneck supported by the task diagnosis. Use prior practice windows, interaction costs, and changes in full-task success to inform the selection when this history is available.
4. Set max_steps to an integer from 50 to 1000 in multiples of 50. Choose a budget appropriate for the selected window and the available evidence.
5. Provide a concise reason linking the selected window to the observed bottleneck and relevant intervention history. Explicitly state uncertainty when the evidence does not support a clear choice.
6. Populate milestones according to the milestone specification below. Use an empty array when no milestone proposal is supported.

Output format requirement: Return only a valid JSON object matching the structure below. Replace task_name with the exact eligible task name and repeat the entry for every eligible task. The values shown are examples. Do not include Markdown fences, comments, or text outside the JSON object.
{
  "windows": {
    "task_name": {
      "start_stage": 0,
      "end_stage": 1,
      "max_steps": 500,
      "reason": "selection reason",
      "milestones": []
    }
  }
}
\end{promptbox}

\begin{promptbox}[breakable,lines before break=5,title after break={Milestone Specification Prompt (continued)}]{Milestone Specification Prompt}
As part of the joint window-selection response, propose fine-grained milestones within each selected window's target subtask. Use the supplied task checker, stage definitions, and supporting evidence to specify how each milestone could be verified.

Requirements:
1. Propose at most four milestones per window. For each milestone, provide a name, completion condition, required simulator signals, source evidence, and missing information. Use an empty array when no milestone is supported.
2. Treat the supplied task checker and stage definitions as evidence. Their presence does not establish that a proposed milestone has already been implemented.
3. Use exact signal or function names only when they appear in the input. Otherwise, describe the required observable conceptually and identify the missing implementation.
4. Do not invent API names, thresholds, or reward weights. Do not interpret missing evidence as failure.
5. Specify each milestone as a one-time reward event upon a new valid completion. Account for the restored entry state and exclude milestones already satisfied at restoration. A condition remaining true must not produce repeated rewards.
6. Across Normal, Prefix, and Suffix Window rollouts, verified milestone rewards replace DAG-based Progress Reward within the selected window. Outside the window, retain DAG-based Progress Reward. Retain the full-task success reward only when the environment confirms success. Normal-start fallback rollouts use the original task reward.

Output format requirement: Populate each window's milestones array using the structure below, with one object per supported milestone. All fields are strings except required_simulator_signals, which is an array of strings. Return the array within the complete joint window-selection JSON object, not as a separate response. Do not include Markdown fences, comments, or text outside that JSON object.
[
  {
    "name": "milestone",
    "completion_condition": "condition",
    "required_simulator_signals": ["signal"],
    "source_evidence": "input evidence",
    "missing_information": "missing details"
  }
]
\end{promptbox}

\section{Experimental Details}
\label{sec:supp_seed_variance}

In this section, we describe the training configurations (\textbf{\S}\ref{sec:supp_implementation}), the data sampling protocols (\textbf{\S}\ref{sec:supp_data_settings}), and the multi-seed training and evaluation settings (\textbf{\S}\ref{sec:supp_multiseed_training_and_evaluation}). We then present ablation studies on window selection across rounds (\textbf{\S}\ref{sec:supp_window_history_ablation}) and the choice of Scheduler model (\textbf{\S}\ref{sec:supp_scheduler_model_comparison}), followed by analyses of reward allocation (\textbf{\S}\ref{sec:supp_reward_allocation_control}) and practice allocation (\textbf{\S}\ref{sec:supp_practice_allocation}).

\subsection{Training Configurations}
\label{sec:supp_implementation}
\label{sec:supp_advantage_ppo}
\label{sec:supp_hardware}

\noindent
\begin{minipage}[t]{0.485\linewidth}
\vspace{0pt}
\noindent \textbf{Hyperparameters.} Table~\ref{tab:supp_robocasa_hparams} summarizes the training configurations. We initialize each policy from its corresponding $\pi_{0.5}$ SFT checkpoint and train it with distributed PPO. We use AdamW with learning rates of $5\times10^{-6}$ for the policy and $10^{-4}$ for the value head, $(\beta_1,\beta_2)=(0.9,0.95)$, and a weight decay of 0.01. We clip the gradient norm at 1.0. The global batch of 512 is split into per-GPU microbatches. Each run comprises three rounds of 25 Global RL iterations followed by 25 Local RL iterations, for 150 PPO iterations in total. In each iteration, we collect fresh trajectories over the specified number of rollout epochs and then make the specified number of PPO passes over the resulting rollout batch.
\end{minipage}\hfill
\begin{minipage}[t]{0.485\linewidth}
\vspace{0pt}
\captionsetup{type=table,aboveskip=0pt}
\centering
\footnotesize
\setlength{\tabcolsep}{2pt}
\renewcommand{\arraystretch}{0.98}
\caption{\textbf{Hyperparameter settings of ReFocus.} Configurations are shown for RoboCasa365 and LIBERO-Long.}
\label{tab:supp_robocasa_hparams}
\begin{tabular*}{\columnwidth}{@{\extracolsep{\fill}}lcc@{}}
\toprule
Benchmark & RoboCasa365 & LIBERO-Long \\
\midrule
GPUs / parallel train envs & 16 / 256 & 8 / 64 \\
Global / GPU microbatch & 512 / 32 & 512 / 32 \\
Rollout / PPO epochs & 2 / 1 & 8 / 4 \\
Action chunk / flow steps & 50 / 10 & 10 / 10 \\
Max. horizon (steps) & 2,900 & 1,024 \\
Actor / critic LR & $5\!\times\!10^{-6}$ / $10^{-4}$ & $5\!\times\!10^{-6}$ / $10^{-4}$ \\
Adam $(\beta_1,\beta_2)$ & $(0.9,0.95)$ & $(0.9,0.95)$ \\
Weight decay / grad. clip & 0.01 / 1.0 & 0.01 / 1.0 \\
Discount $\gamma$ / GAE $\lambda$ & 0.99 / 0.95 & 0.99 / 0.95 \\
PPO policy / value clip & 0.2 / 0.2 & 0.3 / 0.2 \\
KL coefficient & 0 & 0 \\
Temperature (train / eval) & 1.0 / 0.6 & 1.0 / 0.6 \\
\bottomrule
\end{tabular*}

\end{minipage}
\par\medskip

\noindent\textbf{PPO for Flow-Based Policies.}
At environment decision step $k$, $\pi_{0.5}$ generates an action chunk by integrating
$d\mathbf{x}_\tau=\mathbf{v}_\theta\,d\tau$ from noise $\mathbf{x}_1\sim\mathcal{N}(0,I)$
to the clean chunk $\mathbf{x}_0$. Here, $\mathbf{v}_\theta=v_\theta(\mathbf{x}_\tau,\tau,o_k)$
is the velocity field conditioned on observation $o_k$. We follow Flow-SDE~\cite{chen2025pirl,zang2026rlinfvla}. On the linear flow-matching path
$\mathbf{x}_\tau=(1-\tau)\mathbf{x}_0+\tau\boldsymbol{\xi}$,
with $\boldsymbol{\xi}\sim\mathcal{N}(0,I)$ independent of the clean chunk,
the ideal velocity is the conditional mean of $\boldsymbol{\xi}-\mathbf{x}_0$.
Substituting the path equation gives
$\mathbb{E}[\boldsymbol{\xi}\mid\mathbf{x}_\tau,o_k]
=\mathbf{x}_\tau+(1-\tau)\mathbf{v}_\theta$ at the flow-matching optimum.
Since the score of the Gaussian conditional on $\mathbf{x}_0$ is
$-\boldsymbol{\xi}/\tau$, averaging over $\mathbf{x}_0$ given $\mathbf{x}_\tau,o_k$
gives the score approximation:
\begin{equation}
\nabla_{\mathbf{x}_\tau}\log p_\tau(\mathbf{x}_\tau\mid o_k)
\approx-\frac{\mathbf{x}_\tau+(1-\tau)\mathbf{v}_\theta}{\tau}.
\label{eq:supp_flow_score}
\end{equation}
For decreasing $\tau$, the reverse-time SDE has probability-current velocity
$\mathbf{b}_\theta+\tfrac{1}{2}\sigma_\tau^2\nabla_{\mathbf{x}_\tau}\log p_\tau$.
Matching it to $\mathbf{v}_\theta$ and substituting Eq.~\eqref{eq:supp_flow_score} yields:
\begin{equation}
d\mathbf{x}_\tau=\mathbf{b}_\theta\,d\tau+\sigma_\tau\,d\bar{\mathbf{W}}_\tau,\quad
\mathbf{b}_\theta=\mathbf{v}_\theta+
\frac{\sigma_\tau^2}{2\tau}\bigl(\mathbf{x}_\tau+(1-\tau)\mathbf{v}_\theta\bigr),
\label{eq:supp_flow_sde}
\end{equation}
where $\bar{\mathbf{W}}_\tau$ is a reverse-time Wiener process and $\sigma_\tau$ is the diffusion coefficient, scaled in proportion to $\sqrt{\tau/(1-\tau)}$ at interior denoising times; at $\tau=1$, the next grid point is used in the denominator to keep the scale finite.

For a backward step of size $\Delta>0$, Euler--Maruyama gives:
\begin{equation}
\mathbf{x}_{\tau-\Delta}
=\underbrace{\mathbf{x}_\tau-\Delta\mathbf{b}_\theta}_{\text{corrected mean}}
+\underbrace{\sigma_\tau\sqrt{\Delta}\,\boldsymbol{\zeta}}_{\text{Gaussian perturbation}},
\label{eq:supp_flow_step}
\end{equation}
where $\boldsymbol{\zeta}\sim\mathcal{N}(0,I)$ is an independent Gaussian noise sample.
During training, one randomly selected denoising step uses Eq.~\eqref{eq:supp_flow_step}.
The other steps use the ODE update $\mathbf{x}_{\tau-\Delta}=\mathbf{x}_\tau-\Delta\mathbf{v}_\theta$.
Evaluation uses the ODE update throughout denoising.

\noindent\textbf{Likelihood ratio and advantages.}
Equation~\eqref{eq:supp_flow_step} gives a Gaussian transition with mean
$\mathbf{x}_\tau-\Delta\mathbf{b}_\theta$ and covariance $\sigma_\tau^2\Delta I$.
For chunk $k$, we save the selected denoising index, the latents before and after the stochastic step,
and the behavior log-probability $\ell_k(\theta_{\mathrm{old}})$ summed over the executed action coordinates.
We evaluate the same recorded transition under the current actor to obtain $\ell_k(\theta)$.
The ratio in the main-text PPO objective, Eq.~\eqref{eq:repractice_ppo}, is then:
\begin{equation}
\rho_k(\theta)=\exp\!\left(\ell_k(\theta)-\ell_k(\theta_{\mathrm{old}})\right).
\label{eq:supp_flow_ratio}
\end{equation}

Both phases use this ratio. The reward $r_k$ is $r_k^{\mathrm G}$ from
Eq.~\eqref{eq:repractice_reward} in Global RL and $r_k^{\mathrm L}$ from
Eq.~\eqref{eq:local_reward} in Local RL. These rewards enter GAE through:
\begin{equation}
\delta_k=r_k+\gamma V_{k+1}-V_k,\quad
\hat A_k=\delta_k+\gamma\lambda\hat A_{k+1},
\label{eq:supp_phase_gae}
\end{equation}
where $V_k$ is the rollout critic prediction and $\hat{A}_k$ is the GAE advantage estimate. The recursion stops at episode or truncation boundaries.
The next-state value is zero at a true terminal state, and at a truncation it is bootstrapped from
the final observation. For fixed rollout values, a reward change at a later chunk $j$ contributes
to $\hat A_k$ with weight $(\gamma\lambda)^{j-k}$ within the same trajectory.
Thus, the reward definitions determine how completion events contribute to the chunk-level advantages.
The actor uses the normalized advantage $\widetilde A_k$ in Eq.~\eqref{eq:repractice_ppo},
and the critic regresses to the fixed target $V_k+\hat A_k$.

\begin{figure}[t]
    \centering
    \includegraphics[width=\textwidth]{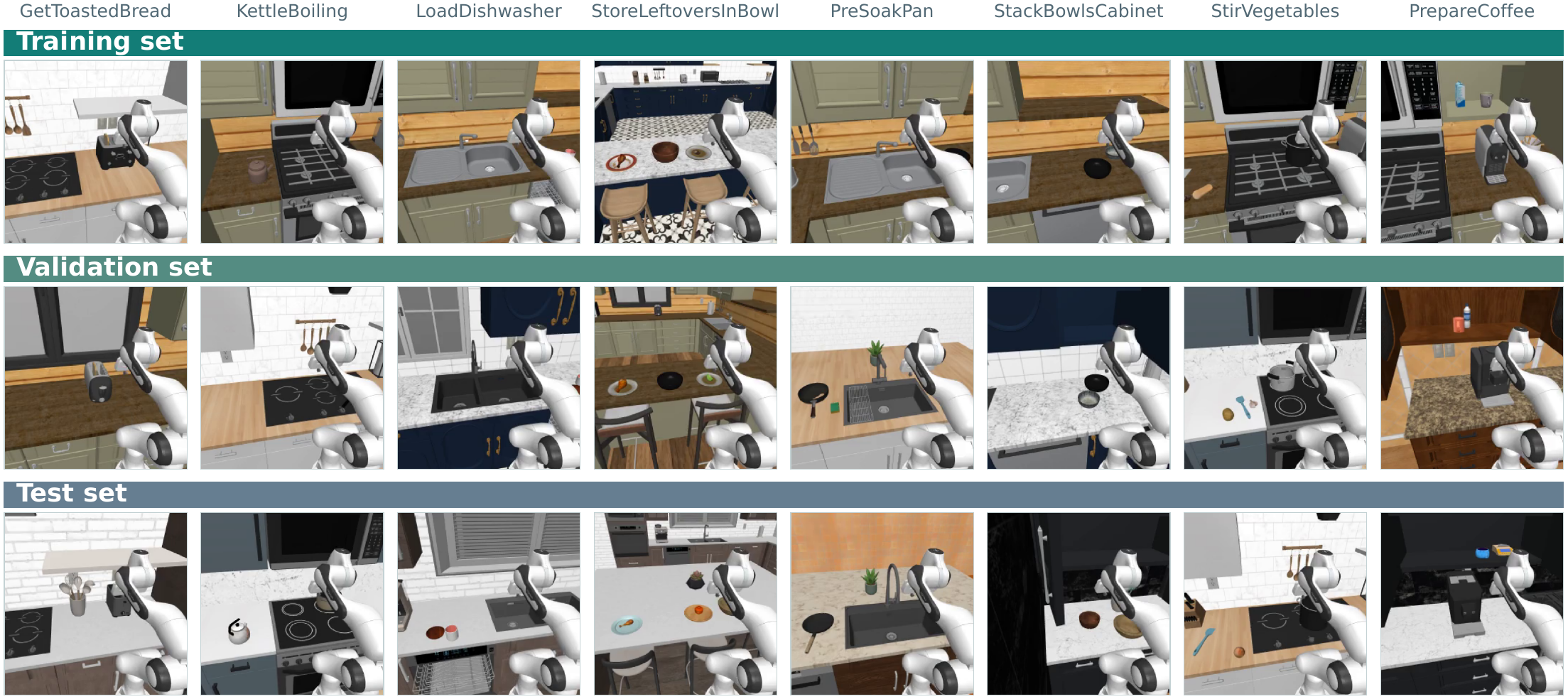}
    \caption{\textbf{Representative RoboCasa365 scenes for training, validation, and testing.}}
    \label{fig:robocasa_scene_sampling}
\end{figure}

\subsection{Data Sampling and Evaluation}
\label{sec:supp_data_settings}

For LIBERO-Long, we sample initial states from the benchmark-provided state bank for each task. For RoboCasa365, we use the \texttt{target} configuration with target object instances and ten paired kitchen layout/style settings, $(1,1),\ldots,(10,10)$. Training rollouts support policy optimization and provide restorable entries, validation rollouts guide bottleneck diagnosis and practice-window selection, and test rollouts measure full-task performance. Figure~\ref{fig:robocasa_scene_sampling} illustrates these three roles on RoboCasa365. The actor uses random seed 1234. RoboCasa365 training and validation environments use base seed 42 with task-index and replica offsets, whereas final testing uses base seed 7. For LIBERO-Long, the environment seed is set to 7, and the 50 benchmark-provided initial states per task, indexed from 0 to 49, are loaded sequentially.

\subsection{Multi-seed Training and Evaluation}
\label{sec:supp_multiseed_training_and_evaluation}

On LIBERO-Long, Table~\ref{tab:seed_std} illustrates variability across three independent training seeds under a common evaluation protocol. Each policy is evaluated on the same 50 benchmark initial states per task. On RoboCasa365, Table~\ref{tab:seed_std_robocasa} summarizes variability across three evaluation seeds under the target protocol while holding each method's checkpoint fixed. For each horizon group and overall, we average task success rates within each seed and report the mean success rate (\%) and standard deviation (percentage points) across seeds.

\begin{table}[!htbp]
\centering
\begin{minipage}[t]{0.485\linewidth}
\vspace{0pt}
\centering
\footnotesize
\caption{\textbf{Multi-seed training on LIBERO-Long.} We conduct three individual runs with different seeds and report mean success rate (\%) with standard deviation ($\pm$).}
\label{tab:seed_std}
\setlength{\tabcolsep}{2pt}
\renewcommand{\arraystretch}{1.05}
\begin{tabular*}{\columnwidth}{@{\extracolsep{\fill}}lcccc@{}}
\toprule
\multirow{2}{*}{Method} & \multicolumn{3}{c}{Horizon Steps} & \multirow{2}{*}{Overall} \\
\cmidrule(lr){2-4}
 & 250--340 & 340--400 & 400--550 & \\
\midrule
$\pi_{0.5}$ & \(93.3\,{\scriptstyle\pm 0.8}\) & \(95.1\,{\scriptstyle\pm 1.0}\) & \(72.3\,{\scriptstyle\pm 1.5}\) & \(89.7\,{\scriptstyle\pm 1.0}\) \\
Sparse-RL & \(96.7\,{\scriptstyle\pm 0.8}\) & \(93.8\,{\scriptstyle\pm 0.8}\) & \(75.3\,{\scriptstyle\pm 1.5}\) & \(91.5\,{\scriptstyle\pm 0.9}\) \\
SimpleVLA-RL & \(96.7\,{\scriptstyle\pm 0.6}\) & \(95.6\,{\scriptstyle\pm 1.0}\) & \(80.3\,{\scriptstyle\pm 1.5}\) & \(93.1\,{\scriptstyle\pm 0.8}\) \\
StructRL & \(97.7\,{\scriptstyle\pm 0.5}\) & \(96.7\,{\scriptstyle\pm 1.2}\) & \(88.3\,{\scriptstyle\pm 0.6}\) & \(95.5\,{\scriptstyle\pm 0.6}\) \\
ReFocus & \(\mathbf{98.9}\,{\scriptstyle\pm 0.5}\) & \(\mathbf{97.8}\,{\scriptstyle\pm 1.0}\) & \(\mathbf{92.3}\,{\scriptstyle\pm 0.6}\) & \(\mathbf{97.3}\,{\scriptstyle\pm 0.6}\) \\
\bottomrule
\end{tabular*}

\end{minipage}\hfill
\begin{minipage}[t]{0.485\linewidth}
\vspace{0pt}
\centering
\footnotesize
\caption{\textbf{Multi-seed evaluation on RoboCasa365 Composite-Seen.} We conduct three individual evaluations with different seeds and report mean success rate (\%) with standard deviation ($\pm$).}
\label{tab:seed_std_robocasa}
\setlength{\tabcolsep}{2pt}
\renewcommand{\arraystretch}{1.05}
\begin{tabular*}{\columnwidth}{@{\extracolsep{\fill}}lcccc@{}}
\toprule
\multirow{2}{*}{Method} & \multicolumn{3}{c}{Horizon Steps} & \multirow{2}{*}{Overall} \\
\cmidrule(lr){2-4}
 & 800--1000 & 1000--1400 & 1400--2900 & \\
\midrule
$\pi_{0.5}$ & \(41.4\,{\scriptstyle\pm 5.0}\) & \(47.0\,{\scriptstyle\pm 1.1}\) & \(32.8\,{\scriptstyle\pm 1.2}\) & \(38.8\,{\scriptstyle\pm 1.1}\) \\
Sparse-RL & \(43.3\,{\scriptstyle\pm 1.1}\) & \(44.0\,{\scriptstyle\pm 2.2}\) & \(34.4\,{\scriptstyle\pm 0.9}\) & \(39.0\,{\scriptstyle\pm 0.9}\) \\
SimpleVLA-RL & \(47.0\,{\scriptstyle\pm 1.0}\) & \(47.4\,{\scriptstyle\pm 0.8}\) & \(35.0\,{\scriptstyle\pm 0.7}\) & \(41.1\,{\scriptstyle\pm 0.4}\) \\
StructRL & \(52.3\,{\scriptstyle\pm 2.6}\) & \(51.3\,{\scriptstyle\pm 0.3}\) & \(38.0\,{\scriptstyle\pm 2.2}\) & \(44.8\,{\scriptstyle\pm 1.1}\) \\
ReFocus & \(\mathbf{58.6}\,{\scriptstyle\pm 1.3}\) & \(\mathbf{53.5}\,{\scriptstyle\pm 0.8}\) & \(\mathbf{45.6}\,{\scriptstyle\pm 1.0}\) & \(\mathbf{50.4}\,{\scriptstyle\pm 1.0}\) \\
\bottomrule
\end{tabular*}

\end{minipage}
\end{table}

\subsection{Ablation Study on Window Selection across Rounds}
\label{sec:supp_window_history_ablation}

\begin{table}[t]
\centering
\begin{minipage}[t]{0.48\textwidth}
\newcommand{\WindowAblationCheck}{{\usefont{U}{pzd}{m}{n}\symbol{51}}}
\newcommand{\WindowAblationCross}{{\usefont{U}{pzd}{m}{n}\symbol{55}}}
\centering
\footnotesize
\setlength{\tabcolsep}{3pt}
\renewcommand{\arraystretch}{1}
\caption{\textbf{Ablation of practice-window selection on LIBERO-Long.} Success rates (\%) are grouped by task horizon.}
\label{tab:ablation_window_history}
\resizebox{\linewidth}{!}{%
\begin{tabular}{@{}lcccccc@{}}
\toprule
\multirow{2}{*}{Method}
& \multirow{2}{*}{Adaptive}
& \multirow{2}{*}{History}
& \multicolumn{4}{c}{Horizon Steps} \\
\cmidrule(lr){4-7}
& & & 250--340 & 340--400 & 400--550 & Overall \\
\midrule
Fixed Window & \WindowAblationCross & \WindowAblationCross & 96.8 & 96.0 & 88.0 & 94.8 \\
Earliest-Failure Rule & \WindowAblationCheck & \WindowAblationCross & 97.6 & 96.7 & 89.0 & 95.6 \\
\midrule
\multirow{2}{*}{LLM-as-Scheduler} & \WindowAblationCheck & \WindowAblationCross & 98.8 & 98.0 & 92.0 & 97.2 \\
& \WindowAblationCheck & \WindowAblationCheck & \textbf{99.2} & \textbf{98.7} & \textbf{93.0} & \textbf{97.8} \\
\bottomrule
\end{tabular}%
}

\end{minipage}\hfill
\begin{minipage}[t]{0.48\textwidth}
\centering
\footnotesize
\setlength{\tabcolsep}{4pt}
\renewcommand{\arraystretch}{1.05}

\caption{\textbf{Scheduler ablation on RoboCasa365 Composite-Seen.}
Success rates (\%) by task horizon.}
\label{tab:ablation_supervisor}

\resizebox{\linewidth}{!}{%
\begin{tabular}{@{}lccccc@{}}
\toprule
\multicolumn{1}{c}{\multirow[c]{2}{*}{Scheduler}}
& \multicolumn{1}{c}{\multirow[c]{2}{*}{Cost / Task}}
& \multicolumn{4}{c}{Horizon Steps} \\
\cmidrule(lr){3-6}
& & 800--1000 & 1000--1400 & 1400--2900 & Overall \\
\midrule
GPT-6 Astra~\cite{openai2026gpt6astra} & \$1.28 & \textbf{59.7} & \textbf{54.2} & \textbf{46.4} & \textbf{51.3} \\
Opus 5.5~\cite{anthropic2026opus55}    & \$0.51 & 58.3   & 52.4   & 44.0  & 49.3  \\
Fable 5.1~\cite{anthropic2026fable51}   & \$1.28 & 59.3  & 53.4  & 45.8   & 50.7   \\
Qwen3.8-27B~\cite{qwen2026qwen3827b} & \$0.05 & 56.0   & 52.6   & 43.8   & 48.8   \\
\bottomrule
\end{tabular}%
}

\end{minipage}
\vspace{-10pt}
\end{table}

Table~\ref{tab:ablation_window_history} compares window selection strategies and the use of the History Buffer on LIBERO-Long. The controls in Table~\ref{tab:ablation_window_history} isolate window selection and prior practice history on LIBERO-Long. Fixed Window retains each task's first-round window bounds and interaction budget throughout training, while refreshing restorable entries from the current round. Earliest-Failure Rule selects a window using the earliest stage at which validation rollouts repeatedly stall, without LLM-based window selection or practice history. Adaptive w/o History reselects windows using current-round diagnoses and available entries, without preceding-round practice records. Adaptive w/ History additionally uses the History Buffer to retain, revise, or skip windows based on previous practice and its outcomes. All variants use three rounds, each with 25 Global RL updates and 25 Local RL updates, and share the SFT initialization, task DAGs, reward definitions, milestone checkers and weights, rollout-mode probabilities, and validation schedule. The same entry eligibility and window budget constraints apply across variants. Milestone definitions and weights for each target remain fixed across rounds. Missing valid entries trigger normal-start fallback. Interaction costs distinguish fresh-interaction steps from prefix replay, as matching PPO updates does not imply equal simulator cost. All variants use the same full-task evaluation protocol.

\subsection{Ablation Study on the Scheduler}
\label{sec:supp_scheduler_model_comparison}

Table~\ref{tab:ablation_supervisor} compares Scheduler models for bottleneck diagnosis and window selection on RoboCasa365 Composite-Seen, reporting success rates by task horizon and per-task inference cost. GPT-6 Astra achieves the highest overall success rate (51.3\%) at \$1.28 per task. Fable~5.1 matches this cost but trails by only 0.6 points overall (50.7\%), making it a competitive alternative. Opus~5.5 reduces cost to \$0.51 while reaching 49.3\%. The open-source Qwen3.8-27B achieves 48.8\% at \$0.05 per task, only 2.5 points below Astra, suggesting that effective bottleneck diagnosis does not require the most capable frontier model. 

\subsection{Reward Allocation Analysis}
\label{sec:supp_reward_allocation_control}

We select three multi-stage RoboCasa365 tasks to examine how the distribution of subtasks affects reward allocation across stages. Stage~1 contains two subtasks in each task. In \texttt{StoreLeftoversInBowl}, the robot puts chicken and vegetables into the bowl. In \texttt{LoadDishwasher}, it loads two dishes. In \texttt{StirVegetables}, it adds two vegetables to the pot. Each later stage has one subtask. With uniform per-subtask rewards, Stage~1 receives 50\% of the progress budget in the first two tasks and 40\% in \texttt{StirVegetables}, while each later stage receives only 25\% or 20\%, respectively. This imbalance means the policy receives a stronger training signal for early-stage completions and a weaker signal for the stages that remain difficult — precisely where additional practice is most needed. Stage-balanced rewards give each stage an equal share of the budget regardless of how many subtasks it contains, preventing any single stage from dominating the reward signal. Table~\ref{tab:ablation_reward_allocation_control} summarizes progress in archived normal-start rollouts. For \texttt{StirVegetables}, 21 of 25 rollouts completed Stage~1, but only 10 completed the final tracked stage, illustrating how uniform per-subtask rewards can allocate more credit to an early stage even when progress frequently stalls later. The reward-feedback ablation on LIBERO-Long (Table~4 in the main paper) further corroborates this: stage-balanced allocation improves overall success by 2.6 points over structured per-subtask rewards, with the gap widening to 9.0 points on the hardest task-horizon group (400--550 steps).

\begin{table}[ht]
    \centering
    \small
    \setlength{\tabcolsep}{4pt}
    \caption{\textbf{Reward allocation in multi-stage long-horizon tasks.}
    Each task has five normal-start rollouts at each of five checkpoints from one multi-task Global RL rollout.
    The stage columns count completion of the first stages, final stages and full-task successes.}
    \label{tab:ablation_reward_allocation_control}
    \begin{tabular}{@{}lccc@{}}
        \toprule
        Task & Stage 1 Complete & Final Stage Complete & Full-task Success \\
        \midrule
        \texttt{StoreLeftoversInBowl} & 24/25 & 17/25 & 16/25 \\
        \texttt{LoadDishwasher} & 24/25 & 20/25 & 17/25 \\
        \texttt{StirVegetables} & 21/25 & 10/25 & 10/25 \\
        \bottomrule
    \end{tabular}
\end{table}

\subsection{Practice Allocation Analysis}
\label{sec:supp_practice_allocation}

Figure~\ref{fig:robocasa_practice_allocation_selected} shows the eight tasks with the most stages and longest horizons among the 16 RoboCasa365 Composite-Seen tasks, where the effect of practice reallocation is most pronounced. During Global RL, training steps are distributed across all stages according to rollout outcomes, leaving bottleneck stages underrepresented relative to their difficulty. Local RL corrects this by restoring entries at the selected stage, so the policy spends a larger fraction of its budget on the stages that need it most. The share of \texttt{LoadDishwasher} rollouts reaching the target stage rose from 88.0\% to 98.7\%, and the share of steps in the target stage rose from 50.0\% to 80.1\% for \texttt{StoreLeftoversInBowl}. This reallocation of practice budget toward identified bottlenecks is what drives the per-round improvement shown in the main results.

\begingroup
\setlength{\intextsep}{6pt}
\begin{figure}[!ht]
    \centering
    \includegraphics[width=\linewidth]{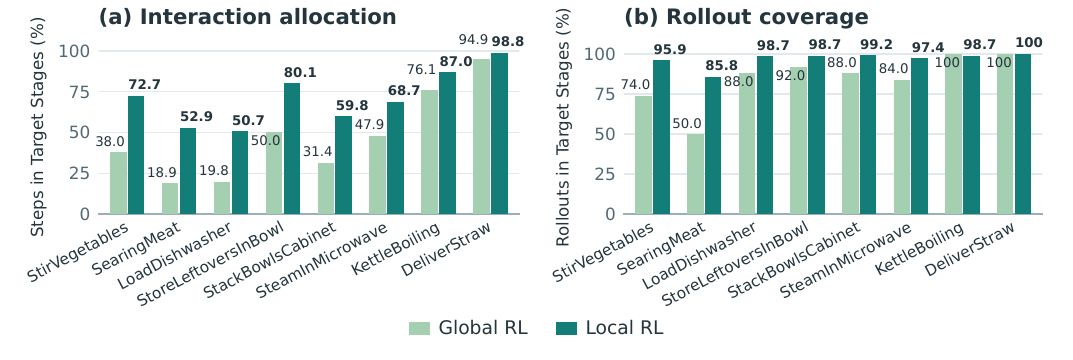}
    \caption{\textbf{Practice allocation in eight RoboCasa365 tasks.} We analyze the second round of a ReFocus run using the same Scheduler-selected stages for both phases. Panel (a) reports the percentage of training steps in those stages. Panel (b) reports the percentage of rollouts with at least one step in those stages.}
    \label{fig:robocasa_practice_allocation_selected}
\end{figure}

\endgroup

\clearpage
\section{Real-World Experiments}
\label{sec:real_world_adaptation}

In this section, we describe the real-world task procedures and trial protocol (\textbf{\S}\ref{sec:supp_real_tasks}), detail the RECAP-based adaptation (\textbf{\S}\ref{sec:supp_recap_adaptation}), and visualize value predictions for successful and failed \texttt{BreadBaker} rollouts (\textbf{\S}\ref{sec:supp_real_value}).

\subsection{Real-World Task Procedures}
\label{sec:supp_real_setup}
\label{sec:supp_real_tasks}

\begin{figure}[!htbp]
    \centering
    \includegraphics[width=\textwidth,trim=0 155 0 0,clip]{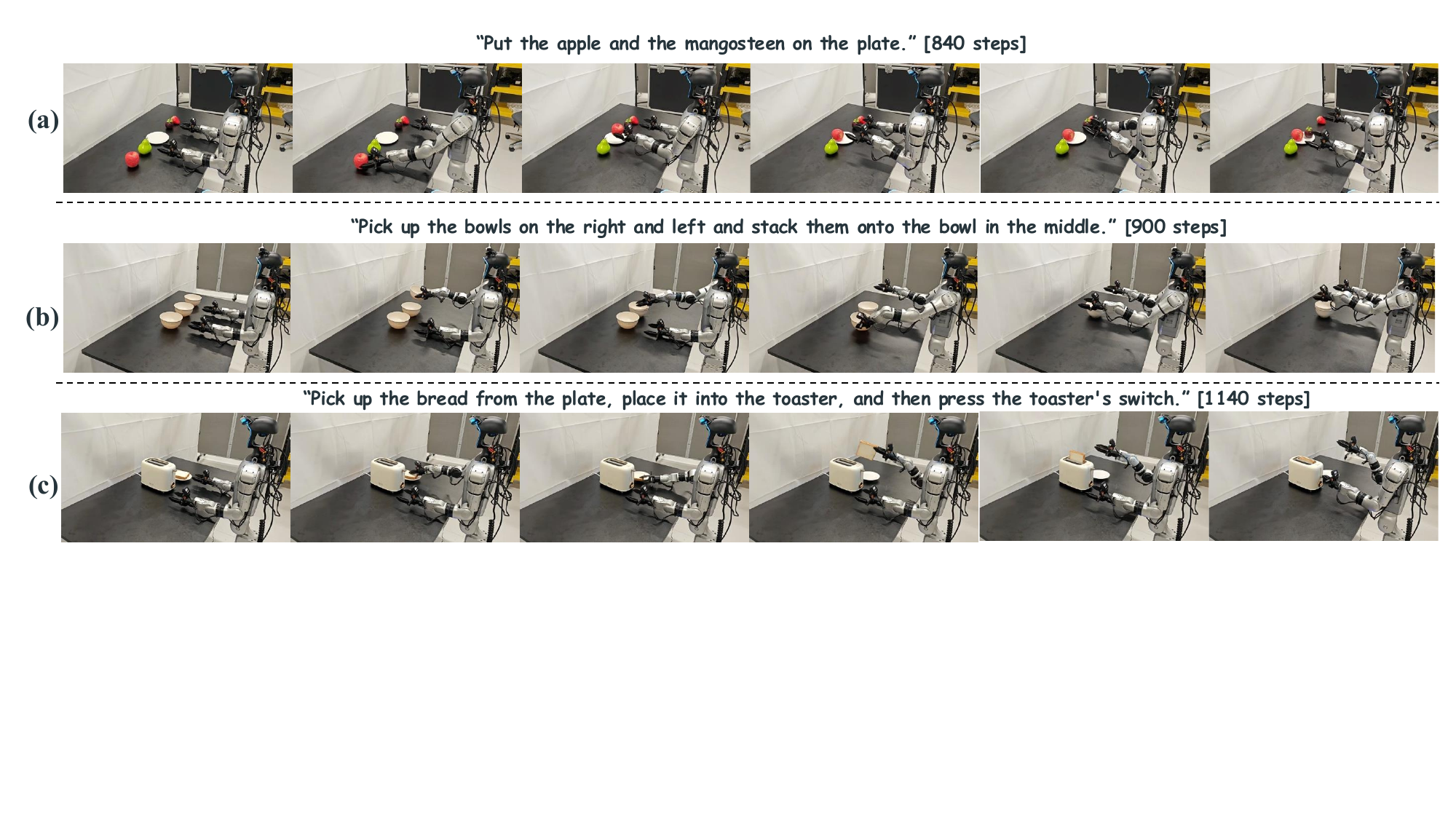}
    \caption{\textbf{Real-world task execution sequences.} Rows (a)--(c) show \texttt{PickFruit}, \texttt{StackBowl}, and \texttt{BreadBaker} , respectively. }
    \label{fig:real_world}
\end{figure}

Figure~\ref{fig:real_world} illustrates the three real-world tasks and their execution sequences. In \texttt{PickFruit}, the robot picks up an apple and a mangosteen and places both on a plate, with 840 steps in the human demonstration rollout. In \texttt{StackBowl}, it picks up the two outer bowls and stacks them onto the middle bowl, with 900 steps in the human demonstration rollout. In \texttt{BreadBaker}, it transfers bread from a plate into a toaster and presses the toaster switch, with 1140 steps in the human demonstration rollout. Each trial starts from the normal task configuration and is limited to 120 seconds for \texttt{PickFruit} and \texttt{StackBowl}, and 180 seconds for \texttt{BreadBaker}.

\subsection{RECAP-Based Adaptation}
\label{sec:supp_recap_adaptation}

We incorporate ReFocus into RECAP~\cite{physicalintelligence2025recap} using autonomous rollouts collected after SFT. We describe the original RECAP formulation and then specify our two modifications.

\noindent\textbf{Background.}
RECAP assigns a per-step reward $r_t^{\mathrm{RECAP}} \in \{-1, 0, -C_{\mathrm{fail}}\}$: $-1$ at every non-terminal step, $0$ on success, and $-C_{\mathrm{fail}}$ on failure. A distributional value model $V_\phi$ is trained by minimizing cross-entropy against empirical Monte Carlo returns $R_t$:
\begin{equation}
\mathcal{L}(\phi) = \mathbb{E}_{\tau \in \mathcal{D}} \left[ \sum_t H\!\left(R_t^B(\tau),\, p_\phi(V \mid \mathbf{o}_t, \ell)\right) \right],
\label{eq:recap_value}
\end{equation}
where $R_t^B$ is the binned return and $H$ is cross-entropy over $B{=}201$ discretized bins. The Monte Carlo return and its binned form are:
\begin{equation}
R_t = \sum_{t'=t}^{T} r_{t'}^{\mathrm{RECAP}}, R_t^B = \mathrm{bin}(R_t),
\label{eq:recap_mc_return}
\end{equation}
where $T$ is the episode length and $\mathrm{bin}(\cdot)$ maps the return to one of $B$ discretized bins. The $n$-step advantage $A_t$ is then binarized into an improvement indicator:
\begin{equation}
I_t = \mathbf{1}\!\left(A_t > \epsilon_\ell\right),
\label{eq:recap_indicator}
\end{equation}
where $\epsilon_\ell$ is set at the 30th percentile of predicted values for task $\ell$. The policy is trained with a negative log-likelihood objective conditioned on $I_t$:
\begin{equation}
\mathcal{L}(\theta) = \mathbb{E}_{\mathcal{D}}\!\left[-\log \pi_\theta(\mathbf{a}_t \mid \mathbf{o}_t, \ell) - \alpha \log \pi_\theta(\mathbf{a}_t \mid I_t, \mathbf{o}_t, \ell)\right].
\label{eq:recap_policy}
\end{equation}
This conditioning on $I_t$ implicitly extracts improved behavior without policy gradient updates on the flow-matching model.

\noindent\textbf{Modification 1: Progress-augmented return.}
RECAP's per-step reward penalizes inefficiency uniformly but provides no signal about which steps made subtask-level progress. We augment it with the stage-balanced progress reward $r_t^{\mathrm{prog}}$ from the DAG-based completion checks:
\begin{equation}
\tilde{r}_t = r_t^{\mathrm{RECAP}} + r_t^{\mathrm{prog}},
\label{eq:recap_return}
\end{equation}
where $r_t^{\mathrm{prog}} > 0$ fires once when a subtask is first completed. Substituting $\tilde{r}_t$ for $r_t^{\mathrm{RECAP}}$ in Eq.~\eqref{eq:recap_value}, the modified value training objective becomes:
\begin{equation}
\tilde{\mathcal{L}}(\phi) = \mathbb{E}_{\tau \in \mathcal{D}} \left[ \sum_t H\!\left(\tilde{R}_t^B(\tau),\, p_\phi(V \mid \mathbf{o}_t, \ell)\right) \right],
\label{eq:recap_value_aug}
\end{equation}
where $\tilde{R}_t^B = \mathrm{bin}(\tilde{R}_t)$ and $\tilde{R}_t = \sum_{t'=t}^{T} \tilde{r}_{t'}$ is the augmented Monte Carlo return. The improvement indicator in Eq.~\eqref{eq:recap_indicator} is then computed with the augmented advantage $\tilde{A}_t$:
\begin{equation}
\tilde{I}_t = \mathbf{1}\!\left(\tilde{A}_t > \epsilon_\ell\right),
\label{eq:recap_indicator_aug}
\end{equation}
where $\tilde{A}_t$ is the $n$-step advantage estimated from $\tilde{R}_t$. Steps where a subtask is completed receive elevated $\tilde{A}_t$, making those segments more likely to be labeled as improvements and reinforced in Eq.~\eqref{eq:recap_policy}. Figure~\ref{fig:recap_value} illustrates this for \texttt{BreadBaker}: the progress-aware value rises during bread insertion and lever operation and drops when the bread returns to the plate, whereas the sparse-only baseline remains flat until episode termination.

\noindent\textbf{Modification 2: Window-guided data resampling.}
Since intermediate simulator states cannot be restored on physical hardware, we replace state-based Local RL with data resampling. The Scheduler selects a practice window $[s_{\mathrm{start}}, s_{\mathrm{end}}]$ from recorded rollouts. We define a reweighted distribution $\tilde{\mathcal{D}}$ that upsamples trajectory segments within the selected window by factor $\alpha > 1$. Substituting $\tilde{\mathcal{D}}$ for $\mathcal{D}$ in Eq.~\eqref{eq:recap_policy} gives the modified policy objective:
\begin{equation}
\tilde{\mathcal{L}}(\theta) = \mathbb{E}_{\tilde{\mathcal{D}}}\!\left[-\log \pi_\theta(\mathbf{a}_t \mid \mathbf{o}_t, \ell) - \alpha_w \log \pi_\theta(\mathbf{a}_t \mid \tilde{I}_t, \mathbf{o}_t, \ell)\right],
\label{eq:recap_policy_aug}
\end{equation}
where $\tilde{\mathcal{D}}$ is the reweighted data distribution and $\alpha_w$ is the advantage-conditioning weight, concentrating the training budget on identified bottleneck stages without any change to the hardware setup or rollout collection procedure.

\subsection{Value Model Visualization}
\label{sec:supp_real_value}

\begin{figure}[!htbp]
    \centering
    \includegraphics[width=\textwidth]{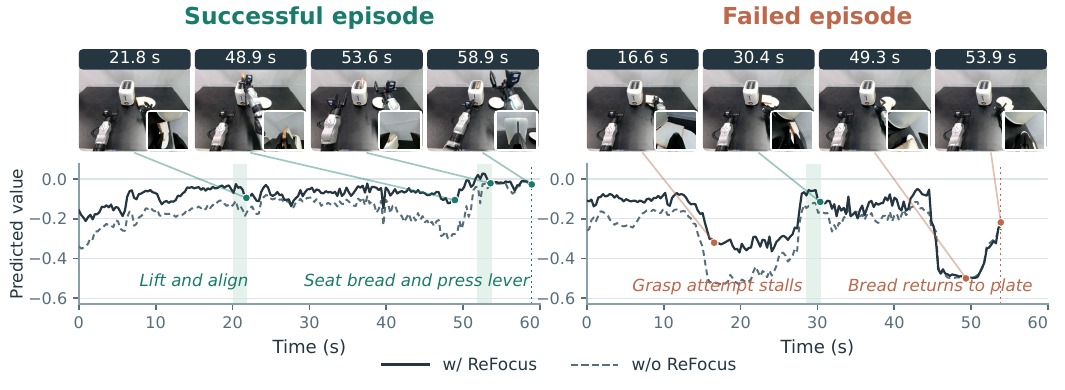}
    \caption{\textbf{Value predictions on real-robot \texttt{BreadBaker} rollouts.} Critics trained with and without progress rewards are evaluated on a successful episode (left) and a failed episode (right). Green and red dots mark progress and failure-related events. }
    \label{fig:recap_value}
\end{figure}

Figure~\ref{fig:recap_value} compares value predictions from critics trained with and without progress rewards on a successful and a failed \texttt{BreadBaker} rollout. The progress-aware critic rises during bread insertion and lever operation, and falls sharply when the bread returns to the plate. In contrast, the sparse-reward baseline remains flat until episode termination.

\par\noindent
\begin{minipage}{\linewidth}
\section{Additional Qualitative Results}
\label{sec:supp_qualitative}

Figure~\ref{fig:supp_libero_qualitative} compares the SFT policy with policies obtained after successive rounds of alternating RL on three LIBERO-Long tasks. All three SFT rollouts shown end in failure. The first successful completions emerge after round~1 for the basket task and after round~2 for the microwave and stove tasks. Among the successful rollouts, the step count at first task completion decreases consistently from round~2 to round~3 across all three tasks, dropping from 378 to 259 steps on \texttt{MugInMicrowave}.

    \centering
    \includegraphics[width=\linewidth]{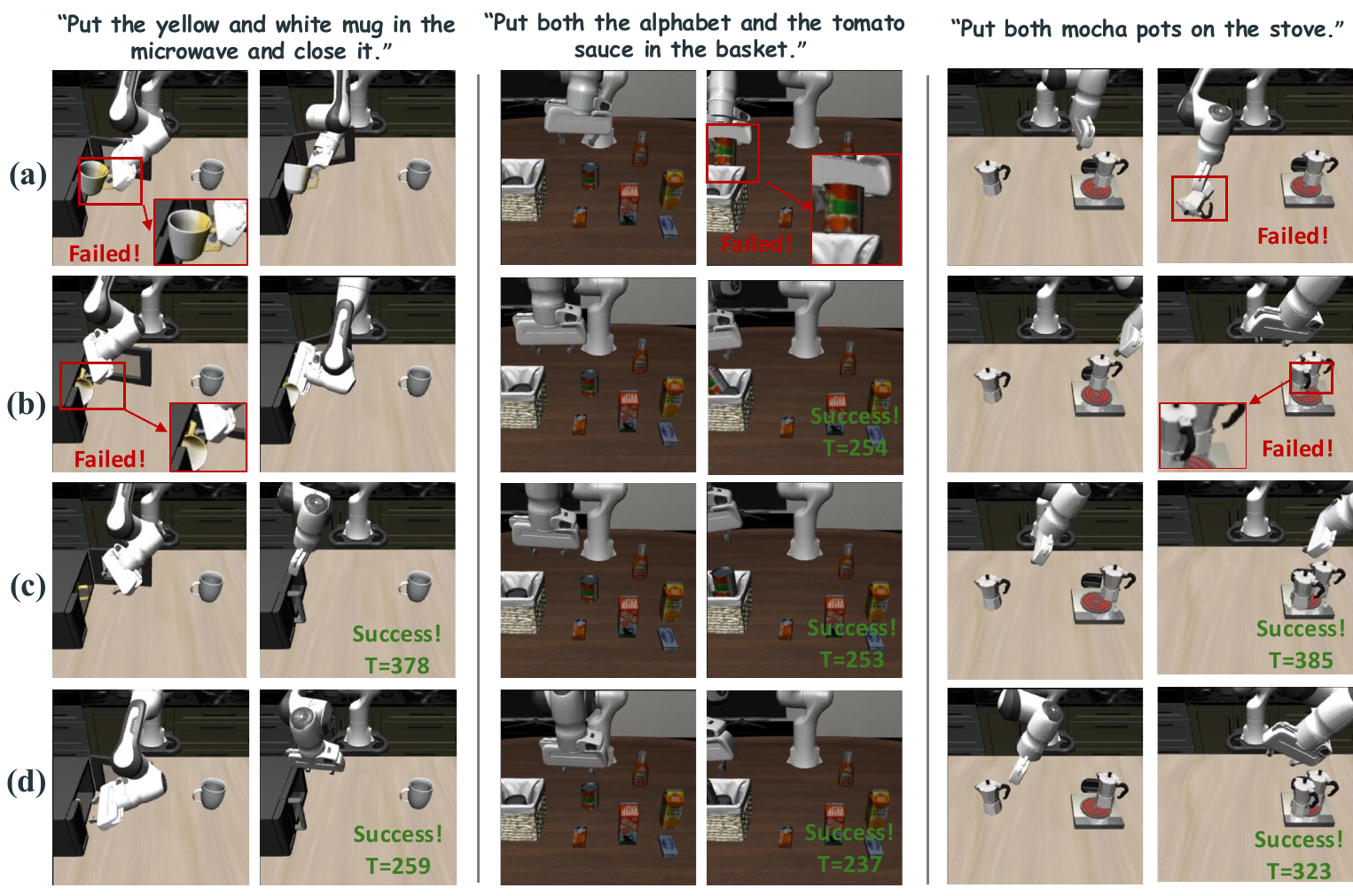}
    \captionsetup{hypcap=false}
\captionof{figure}{\textbf{Qualitative results on three LIBERO-Long tasks.} Rows (a)--(d) show the evaluation results from the SFT policy and alternating RL rounds 1--3. Red boxes highlight failures. $T$ denotes the first successful control step.}
    \label{fig:supp_libero_qualitative}
\end{minipage}
\par

\section{Limitations and Future Work}
\label{sec:supp_limitations}

Some Local RL rollout modes rely on restoring intermediate simulator states. Since comparable mid-episode resets are impractical on physical hardware, the real-world experiments of ReFocus employ RECAP-based post-training on recorded rollouts as an alternative. 

In current version, ReFocus employs a frontier LLM as the Scheduler, which operates between training rounds to diagnose execution bottlenecks and allocate practice windows accordingly. A direction for future work is to extend from an inter-round advisor to an online Co-Worker that remains active throughout task execution. Under this paradigm, when the VLA policy stalls at a bottleneck subtask, the Co-Worker would intervene by selecting and triggering an appropriate recovery skill, temporarily assuming control until task progress is verified, at which point execution would be returned to the policy.

\end{document}